\pdfoutput=1
\documentclass[12pt,twoside,abbrevs,msc,ai,logo,sansheadings]{infthesis}
\usepackage{graphicx}
\usepackage{abbrevs}
\usepackage{acronym}
\usepackage{multirow}
\usepackage{array}
\usepackage{amsmath}
\usepackage{cite}
\usepackage{subfig}
\usepackage{wrapfig}

\title{Parameter Adaptation and Criticality in Particle Swarm Optimization}
\author{Carlos Garc\'ia Cordero}
\submityear{2012}
\date{\today}

\newacro{SQL}{Structured Query Language}
\newacro{HQL}{Hibernate Query Language}
\newacro{ORM}{Object Relational Mapper}
\newacro{DBMS}{Database Management System}
\newacro{FUSE}{Filesystems in Userspace}
\newacro{GUI}{Graphical User Interface}
\newacro{NFS}{Network Filesystem}
\newacro{DSL}{Domain Specific Language}

\abstract{

Generality is one of the main advantages of heuristic algorithms, as such, multiple parameters are exposed to the user with the objective of allowing them to shape the algorithms to their specific needs. Parameter selection, therefore, becomes an intrinsic problem of every heuristic algorithm. Selecting good parameter values relies not only on knowledge related to the problem at hand, but to the algorithms themselves. 

This research explores the usage of self-organized criticality to reduce user interaction in the process of selecting suitable parameters for particle swarm optimization (PSO) heuristics. A particle swarm variant (named Adaptive PSO) with self-organized criticality is developed and benchmarked against the standard PSO. Criticality is observed in the dynamic behaviour of this swarm and excellent results are observed in the long run. In contrast with the standard PSO, the Adaptive PSO does not stagnate at any point in time, balancing the concepts of exploration and exploitation better.

A software platform for experimenting with particle swarms, called PSO Laboratory, is also developed. This software is used to test the standard PSO as well as all other PSO variants developed in the process of creating the Adaptive PSO. As the software is intended to be of aid to future and related research, special attention has been put in the development of a friendly graphical user interface. Particle swarms are executed in real time, allowing users to experiment by changing parameters on-the-fly.

}

\begin{document}
  \begin{preliminary}
    \shieldtype{1}
    \maketitle

    \begin{acknowledgements}

I would like to thank my supervisor, Michael Herrmann, for his insight, intuition and thoughtful e-mails.

A non-quantifiable amount of thanks are extended to my family and friends: To my father, mother and sister for their support. To Chris, Daniel, Danilo, Dino, Nick and all others who endured the harshness of the Scottish life with me. To Natalia, for her lack of circadian rhythms inspired me to finish this work.

My sincere gratitude is extended to all funding bodies that enabled me to pursue my studies: CONACYT (National Science and Technology Council), Santander and the Edinburgh School of Informatics.

\end{acknowledgements}

    \standarddeclaration
    \dedication{

  This is dedicated to you, the reader.

}

    \tableofcontents
    \listoffigures
  \end{preliminary}

  \chapter{Introduction}

Mathematical optimization studies and provides methods for finding the best solution among a set of possible solutions, and albeit this simple definition, optimization is an interesting area of constant and active research. This interest stems from the fact that multitude of problems can be defined as optimization tasks, such as recognizing faces in pictures or hand-written text. The field of machine learning is well known for approaching many different problems in this manner. There are methods, such as logistic regression, which rely on analytical optimization; however, a high percentage of methods used in machine learning (\ie neural networks, SVMs, gaussian processes, etc) rely entirely or partially on search-based iterative methods for optimization \cite{Wright2011}. Developing exact optimization methods is often times difficult and may even be intractable. Search-based methods, in contrast, are easy to implement and are usually the first approach at optimization.

\section{Heuristic Algorithms}

Heuristic algorithms are optimization tools which do not require specialized knowledge from the underlying problem being optimized. According to \cite{Zanakis1981}, because problem-specific knowledge is decoupled from the algorithms themselves, they are excellent candidates for solving optimization tasks where not enough background information is available. This confers the algorithms a certain level of generality which enables them to be useful for solving many different, and often unrelated, optimization problems. Tied to the effectiveness of the algorithms, nevertheless, is effective parameter selection.

Most heuristic algorithms behave according to user supplied parameters. Parameter selection is tied to problem-specific details which in essence decreases the level of generality associated with such algorithms. These parameters are responsible for shaping and making the heuristic algorithms useful in different contexts and situations. Particle swarms, for example, utilize parameters to specify how and where the particles in the swarm explore. These parameters are not trivial to select and the quality of the solution directly depends on these selections. Parameter selection either becomes an interactive task where users tune parameters through trial and error, or knowledge about the problem domain and heuristic algorithm are coupled together to select suitable ones.

\section{Parameter Selection}
\label{sec:parameter_selection}

Several approaches to parameter tuning exist. On the one hand, meta-heuristic algorithms try to find suitable parameters by running a second layer heuristic on top of the desired heuristic. The purpose of this second layer heuristic (or meta-heuristic) is to find a suitable set of parameters where the heuristic being tuned would perform reasonably well. On the other hand, hyper-heuristic algorithms try to find a suitable chain of simple predefined heuristic algorithms that, when executed in a particular order, yield good and acceptable results \cite{Burke2003}. The difference between these two approaches might seem subtle; yet, this is not the case. Meta-heuristic algorithms try to find good parameters for heuristic algorithms targeted at specific problems. Hyper-heuristics do not find parameters. These work with a set of relatively simple heuristics, each having predefined parameters. In this sense, they are targeted towards more general applications where users only need to deploy and forget about them. 
Hyper-heuristics are explained in more detail in section \ref{sec:hyper_heuristics}.

We propose, however, an additional strategy for performing parameter tuning: that which involves an automatic control of parameters without user interaction. In theory, it should be possible to integrate the task of parameter selection into the heuristic algorithms themselves. Just as heuristic algorithms search for suitable solutions among a predefined set of available solutions, parameter selection is analogously another task in this same style: there are specific combination of parameters, among all possible combinations, for which the heuristic algorithm will perform suitably well. How parameter fitness is measured, however, might not be straightforward. To assess the performance of the parameters the question ``Are the parameters working?'' needs to be considered, and assuming that it is possible to answer that question, what follows is ``When should the current parameters be changed in search for better ones?''. These questions are not easy to answer and, in fact, might have different appropriate 
answers rather than a single correct one \cite{El-Gallad2002}. For this reason, we need to turn to the topics of exploration versus exploitation of solutions.

\section{Exploitation Versus Exploration}

The success of any search based optimization algorithm directly depends on the balance of exploitation and exploration of current solutions \cite{Trelea2003}. When a specific set of parameters which seems to outperform all other candidates is repeatedly used throughout different iterations of the heuristic algorithm, we say that the set is being exploited for rewards. In contrast, when the current set of parameters deemed to be the best is put aside to try a new set of unknown parameters, we are performing exploration.

There are multiple models to choose from for performing these exploitative and exploratory tasks; nonetheless, the atypical model of self-organized criticality is examined for reasons that will be made clear in this and subsequent chapters. Typical approaches for preserving diversity in heuristics, and consequently balancing exploitive and exploratory behaviours \cite{Ratnaweera2004}, consist in modifying the heuristics by adding parameters or more complex operations. Arguably, the main advantages of heuristic algorithms lie in their generality and simplicity \cite{Zanakis1981}. Typical approaches either decrease the level of generality by adding more parameters or decrease the level of simplicity by adding complex operations. The approach considered in this project involves using a concept which would neither make the algorithm more complex nor less general: that of self-organized criticality.

\section{Self-organized Criticality}
\label{sec:soc_intro}

Self-organized criticality (SOC) is considered by many a holistic\footnote{Holistic theory states that high level features do not depend on the mechanics of minuscule low level features. Therefore, high level systems should not be explained by means of their low level internals. For example, understanding reasoning as it is accomplished by humans cannot be explained by the low level interaction of neurons in the brain.} approach for explaining complicated dynamic systems as a whole  \cite{Bak1991}. Despite it being a phenomenon commonly observed in nature; from which plenty of heuristic algorithms are inspired, it is not used much in this field. Self-organized criticality possesses some desired characteristics that could enable automatic and effective parameter selection without user interaction in heuristic algorithms.

Critical systems are said to lie at the border of stability and instability. On one side, that of stability, the system tends to converge to a static resting state. On the other side, that of instability, the system tends to diverge into a constant agitated state. Self-organized critical systems are those which are able to return to a critical state on its own without external help after being perturbed by external sources \cite{Bak1991}. The best way to understand criticality is by means of an example. The classical and most typical example of such a system is found in idealized dynamic models of sand piles.

The dynamics of a sand pile are simple in nature. Let us consider the construction of a sand pile from scratch by adding individual sand particles on top of the pile at random locations. If no sand is added, the pile remains static in a stable state. The hight of different locations increase as sand is added into the system. Significant differences in hight will trigger neighbouring particles to rearrange; we call these rearrangements avalanches. There is a point in time (a critical point) where adding just one sand particle will trigger avalanches not only of neighbours, but also of sand throughout the entire pile. The effects of a single sand particle drastically affect the entire system. This is the bordering state between stability and instability. Any disturbance in the system will make the system unstable, but if left as it is, the system will remain stable. 

If we measure the size and frequency of the avalanches occurring throughout the sand pile after a grain of sand is added, the resulting distribution will resemble a power-law \cite{Bak1988}. The interpretation of these results is straightforward: small-sized avalanches occur with much more frequency than big ones. It is in fact believed that any critical system exhibits a power-law distribution explaining some of its dynamics \cite{Bak1991}; the opposite, nonetheless, is not true. The existence of a power-law does not imply an underlying critical system \cite{Beggs2012}.

The addition of criticality to the behaviour of genetic algorithms \cite{Fernandes2008}, differential evolution \cite{Krink2000} and particle swarms \cite{Lovbjerg2002} has proven to be beneficial to them. A heuristic algorithm conferred with the ability to be self-organized and critical with respect to its own parameter selection could behave in a desirable manner. Such systems, in theory, would exploit the best set of parameters with a relatively high frequency without abnegating new possible sets. Furthermore, the exploration capabilities, because of self-organized criticality, would not be hindered by fixed boundaries. Criticality would have the job of determining the boundaries of parameter exploration by increasing or decreasing them with step sizes distributed with a power-law.

This research focuses on the practical aspects of incorporating parameter selection into a heuristic algorithm through the usage of self-organized criticality to drive exploratory and exploitive behaviours in its search patterns. To narrow down the scope of the project, particle swarm optimization algorithms are chosen to be modified given their simplicity, their limited number of parameters, and proven effectiveness in real world applications \cite{Poli2008} \cite{Shi2001}.

Incorporating parameter selection in the particle swarm itself will involve conferring the system with the ability to return to a critical state in the event of perturbation of the parameters. Modifying parameters in the system will greatly affect the particle swarm just as a new grain of sand would cause the whole sand pile to rearrange itself.

\section{A Brief Overview of Particle Swarms}

Particle swarms are the heuristic algorithm of choice of recent years, as demonstrated by the number of published articles covering this topic in comparison to other heuristics \cite{Poli2008} \cite{Selleri2006}. This algorithm was inspired by observing the behaviour of swarms such as birds in a flock or fish schooling. Swarms as a whole perform better than the sum of the individual efforts of its members \cite{Bak1988}. Each member of the swarm interacts with others using only simple and limited actions. Particle swarms usually use three components (or parameters) to specify the searching behaviour of each individual: a social component, a cognitive component and an inertia component. The cognitive component is responsible for weighting the importance an individual gives to its own knowledge of the world. The social component weights the importance an individual gives to the cumulative knowledge of the swarm as a whole. And finally, the inertia component specifies how fast individuals move and change 
direction over time. The topic of particle swarm optimization is covered in more detail in section \ref{sec:pso_background}.

\section{Objectives of the Project}
\label{sec:objectives}

We believe it is possible to automate parameter selection in the particle swarm algorithm by making it critical and self-organized with respect to its behaviour for modifying its own parameters. This could in theory detach the task of selecting suitable parameters from the user. In order to achieve this, two other objectives need to be addressed first. It is required to find a way to incorporate self-organized criticality into the algorithm first, and second, find ways to modify the parameters according to the new behaviour. 

With a new modification of the particle swarm, it will be possible to perform benchmarks and comparisons against the standard particle swarm algorithm using well known problems from the literature.

\section{Project Outline}

In this chapter the topics have only been covered superficially. The next chapters expand and develop each topic more thoroughly and formally. The chapters are structured as follows. \emph{\textbf{Chapter 2}} covers the background material required to better understand the different topics already introduced here. \emph{\textbf{Chapter 3}} summarizes and criticises some of the previous work related to the objectives of this research. The thought process and procedure for designing a self-organized and critical particle swarm algorithm is described in \emph{\textbf{Chapter 4}}. A platform, or software laboratory, was built to test the theory behind the modifications to the particle swarm optimization algorithm. \emph{\textbf{Chapter 5}} describes how the software works and its graphical user interface. \emph{\textbf{Chapter 6}} puts the created software to the test. Several experiments are performed testing and comparing the standard particle swarm and two new modifications. The experimental results are 
presented and analysed. Finally, in \emph{\textbf{Chapter 7}} conclusions are drawn regarding the design, implementation and evaluation performed in this work as well as future paths of related research.

  \chapter{Background}
\label{ch:background}

This chapter covers the background material required to understand the process by which the objectives presented in the introduction are approached. Influential material helping to model the ideas and mechanisms explained and developed in further chapters is also covered. Heuristics and optimization are briefly defined for completeness. The main concepts of hyper-heuristics and meta-heuristics are exposed as they were a helpful insight into the topics of automatic parameter selection. Particle swarms are explained in detail as they represent the core of this research. The trade-off problem of exploration and exploitation is reviewed followed by a detailed explanation of the concepts of self-organized criticality. Finally, an explanation is given of how particle swarms are mixed with self-organized criticality to leverage the exploitation versus exploration trade-off.

\section{Heuristics and Optimization}

Heuristics are non-specialized generalizable procedures for finding approximately optimal solutions to optimization problems. Unconstrained optimization is defined as the $D$-dimensional minimization problem 
\[
\arg\min_{\vec{x}} f(\vec{x}), \; \vec{x} = (x_1, x_2, ..., x_D).
\]
Heuristics approach the problem of minimization (or maximization) with search strategies controlled by used supplied parameters. These problem solving techniques are experience-based and often guided by common sense \cite{Pearl1984}. Their objective is not to find optimal solutions, but rather good enough solutions given the information presently available. More often that not these tools draw their inspiration from nature by trying to simulate how things are perceived to be working. Ant colony optimization uses the well known mechanisms of communication used by ants to guide search. Particle swarms simulate how individuals in a swarm interact using simple rules to create complex systems. Genetic algorithms use mechanics described in evolutionary theory to find good solutions using the concept of survival of the fittest. 

Heuristics used to have and, arguably, still have bad reputation among the academic community. It is often stated that heuristic algorithms are ``the poor man's tool of the trade'' \cite{Eilon1977}. This argument is supported by the fact that heuristics often times lack rigorous mathematical proofs, theorems or convergence demonstrations. Mathematical rigour, nevertheless, does not guarantee success in practice \cite{Bak1991}.

Despite past problems, in the 70s there was a sudden and growing interest in heuristic algorithms \cite{Fisher1980}. The works of R. Karp \cite{Karp1975}, \cite{Karp1975a} and \cite{Karp1976} are believed to have been of great importance to this change \cite{Zanakis1981}. Karp demonstrated the existence of practical combinatorial problems which could not be solved by exact algorithms efficiently neither in time nor in their memory usage. These problems were termed NP-complete. He also created a framework for probabilistic analysis of heuristic algorithms.

The previous paragraph asserts that heuristic algorithms can be taken seriously; still, two important questions are yet to be addressed. ``What are heuristic algorithms good for?'' and ``When should heuristic algorithms be avoided?''. In \cite{Zanakis1981}, practical guidelines are provided regarding these questions. Heuristics should be used when
\begin{enumerate}
 \item Inexact or limited data is available.
 \item An inaccurate, yet rigorous, model is currently being used.
 \item An exact method is not available
 \item An exact method is available but is computationally intractable.
 \item Better performance is required and there is space for approximate rather than optimal solutions.
\end{enumerate}
In general, heuristic algorithms are good when no background information is available, there is not enough data, there is a limited amount of data or when there is complete uncertainty. They also have an important role in multi-objective optimization where exact algorithms are scarce and difficult to design \cite{Fonseca1998}.

In the present, heuristic algorithms are yet again falling in popularity \cite{Hosny2010} as great contributions from the past settle and new research avenues remain dormant. This research focuses on mixing the two previously unrelated concepts of self-organized criticality and parameter selection in heuristic algorithms. We hope this research path  brings forth useful concepts to this field with the goal of pushing progress forward.

\section{Hyper-Heuristics}
\label{sec:hyper_heuristics}

Parameter selection is crucial to the performance of any heuristic algorithm. Section \ref{sec:parameter_selection} introduced several tools for choosing parameters avoiding user interaction as much as possible. These tools helped shape the idea of choosing parameters through adaptation. Hyper-heuristics were also the first ones, along with meta-heuristics, to propose different strategies for dealing with the intricate task of automatically choosing suitable parameters. To a large extent, the main goal of hyper-heuristics, which is also a goal shared in this project, is to raise the level of generality at which optimisation systems operate \cite{Burke2003}. This section covers in more detail hyper-heuristics as they are of great inspirational value to this work. 

Hyper-heuristics do not try to solve optimization problems directly; instead, they try to learn a process for generating good solutions \cite{Ross2002}. A process in this context is defined as a sequence of simple and well understood heuristic algorithms executed in sequence to transform the state of the problem. The order in which the heuristics are applied must be learned and certain heuristics can only work with specially crafted states; these are the only hard implicit constraints in the solution process. Heuristics have the added options of being able to solve the problem directly or model the problem into another state which other heuristics could use.

After an algorithm is applied to a problem, the state of the problem is changed. This is the key concept used in hyper-heuristics. Due to the no-free-lunch theorem\footnote{The ``no-free-lunch'' theorem states that any two search algorithms will have similar performance when averaged across all possible problems. In other words, every algorithm has its weaknesses and strengths which average out \cite{Burke2003}.}, we know there are problems (or states of a problem) for which different heuristics work well. Hyper-heuristics try to select the best heuristic available among a pool of reasonably well known heuristics matching the conditions under which their performance is good \cite{Burke2003}. If several heuristics are combined with this strategy, the most probable worst case performance scenarios of individual heuristics are lost \cite{Ross2002}.

Burke \etal propose a simple, yet straightforward framework for implementing hyper-heuristics \cite{Burke2003}. It consists of 4 steps involving the application of heuristics, the modification of problem states and the finding of good solutions. The framework is summarized as follows:
\begin{enumerate}
\item Create a predefined set of heuristics $\mathcal{H}$, each of which transforms the state of the problem from $\mathcal{S}_t$ to the new state $\mathcal{S}_{t+1}$.
\item Let the initial problem state be defined as $\mathcal{S}_0$.
\item For every state $\mathcal{S}_t$, find the most suitable heuristic in $\mathcal{H}$ for transforming the state.
\item Go to step 3 until the problem is solved or a suitable result is obtained.
\end{enumerate}

Each area of the framework, Burke \etal affirm, has significant opportunities for research. For example, the set of heuristics $\mathcal{H}$ could be evolved throughout the same or different runs of the algorithm. Instead of remaining static, heuristics could modify the parameters of themselves or even others. In the third area of the framework, different algorithms could be run in parallel depending on the state of the problem. This would enable the ranking of heuristics given a particular state. Parallel exploration could potentially create diverse solutions; however, these would be fractured into different algorithms. The ability to merge results from different heuristics would also be required.

Even though standard hyper-heuristics do not perform parameter selection through any means of adaptation, a side effect of having multiple predefined heuristics is that of parameter selection: It is possible to have the initial set of heuristics populated with repeated algorithms instantiated with different parameters or variants. The hyper-heuristic algorithm would then be able to tell, for each state the problem is in, which parameters or variants work best \cite{Burke2003}. 

\section{Particle Swarm Optimization}
\label{sec:pso_background}

Particle swarm heuristic algorithms have been growing continuously in popularity \cite{Spears2010} \cite{Eberhart2001a} since their inception in 1995 \cite{Kennedy1995}. Many variations exist, but in this section only the \textit{original} and the \textit{standard} variations are covered in detail. Other variations relevant to this project are covered in more detail in Chapter \ref{ch:related_work}.

The particle swarm optimization (PSO) algorithm was created as an alternative to standard evolutionary strategies \cite{Kennedy1995}. There are still similarities shared, but the main evolutionary components of mutation, crossover and selection are now missing. PSO was inspired by observing how swarms, such as birds in a flock or fish schooling, behaved in nature; specifically, how the swarm as a whole performs better than the sum of the individual efforts of its members. Every member of the swarm gets to interacts with other members using only simple and limited actions.

A swarm consists of $N$ particles where each particle $i \in  {1, 2, ..., N}$ has a $D$-dimen-sional position and velocity
\begin{align*}
\vec{X_i}(t) &= (x_1(t), x_2(t), ... , x_D(t)) \\
\vec{V_i}(t) &= (v_1(t), v_2(t), ... , v_D(t)).
\end{align*}
Throughout the literature PSOs are represented in several ways using alternative notations. This has spawned several variations of the original standard algorithm according to how each author interprets the notation. The mathematical model of the original PSO is defined as
\begin{align*}
\vec{V}_i(t+1) &{}= \vec{V}_i(t)  + \alpha_1 r_1 (\vec{P_i} - \vec{X_i}(t)) + \alpha_2 r_2 (\vec{G} - \vec{X_i}(t)) \\
\vec{X}_i(t+1) &{}= \vec{X}_i(t) + \vec{V}_i(t+1),
\end{align*}
where $\vec{V}_i(t)$  represents the velocity of particle $i$ at time $t$, and $\vec{X}_i(t)$ represents its position. The velocity of a particle ($\vec{V}_i(t+1)$) is calculated using three components: the inertia ($\vec{V}_i(t)$), cognitive ($\alpha_1 r_1  (\vec{P_i} - \vec{X_i}(t))$), and social ($\alpha_2 r_2  (\vec{G} - \vec{X_i}(t))$) components. The learning rates $\alpha_1$ and $\alpha_2$ control the weight given to the cognitive and social component respectively. The random numbers $r_1$ and $r_2$, in the interval $[0,1]$, are used to make the swarm stochastic. The cognitive component uses the best position so far $\vec{P_i}$ of particle $i$ to modify the current velocity. The social component weights the current velocity by the best position found among all the particles in the swarm, also called ``the global best'' position, $\vec{G}$.

The ambiguity of this definition lies in the random numbers $r_1$ and $r_2$. It is not entirely clear whether they are scalars or vectors, and it is not obvious when they need to be calculated. One possibility is to calculate the random numbers once for each particle, updating every dimension using the same values. The other possibility is to calculate them before updating every dimension of the particles. Despite what looks to be a subtle difference, in \cite{Spears2010} it is pointed out that the first variation is rotationally invariant, while the second is not.

Kennedy, the creator of the PSO algorithm, noticed this and pointed to the second variation as the preferred choice. This variation, he said, had better exploratory properties desirable in a heuristic algorithm \cite{Spears2010}. Because of this, Poli introduced a more accurate notation for the standard PSO \cite{Poli2009}:
\begin{align}
\vec{V}_i(t+1) &{}= \omega \vec{V}_i(t)  + \alpha_1 \vec{r_1} \odot (\vec{P_i} - \vec{X_i}(t)) + \alpha_2 \vec{r_2} \odot (\vec{G} - \vec{X_i}(t)) \label{eq:pso_vel}  \\
\vec{X}_i(t+1) &{}= \vec{X}_i(t) + \vec{V}_i(t+1). \label{eq:pso_pos}
\end{align}
Let us set aside the new coefficient $\omega$ briefly and describe the subtle, and yet important changes in notation. The random numbers $\vec{r_1}$ and $\vec{r_2}$ are now explicit vectors of size $D$, the same dimensions as the particles. The symbol $\odot$ denotes component-wise multiplication. The ambiguities have been removed. Without confusion, each dimension is associated with a different random number now. Notice that the coefficients $\alpha_1$ and $\alpha_2$ remain constant throughout calculations of the velocity. 

$\omega$ is called the inertia coefficient and is used to limit the impact of the previous velocity of the particles. This component was not present in the original definition of the PSO, but quickly became widespread in the literature as the \textit{standard} PSO \cite{Shi1998}. When $\omega$ was first introduced, the parameter remained constant throughout the entire run of the algorithm. Yet, the previous definition of the PSO was modified again into what is now considered the \textit{standard} PSO \cite{Selleri2006}. Instead of being constant, $\omega$ is typically adapted linearly from the predefined constant $\Omega_{top}$ to $\Omega_{bottom}$. As such, in addition to equations (\ref{eq:pso_vel}) and (\ref{eq:pso_pos}), something like the following was introduced:
\begin{align}
\omega(t) = \Omega_{top} - \left\lbrace  \left(\frac{I(t)}{I}\right) \times \left(\Omega_{top} - \Omega_{bottom}\right) \right\rbrace,
\end{align}
where $I(t)$ is the iteration number at time $t$ and $I$ is the total number of iterations to perform.

The standard PSO with constant $\omega$ has been selected to perform adaptive parameter selection using self-organized criticality. Even though there are further modifications to the standard PSO which can improve its performance, such as \cite{Xiang2007}, \cite{Suresh2008}, \cite{Liu2005} and \cite{Hsieh2008}, these variations tend to work better for specific problems by increasing their complexity and adding more parameters. The standard PSO uses a small quantity of parameters ($\omega, \alpha_1, \alpha_2$) and works relatively well all-around. These are the main reasons for this selection.

\section{The Exploration and Exploitation Trade-off}

A trade-off consists of gaining something on one front while loosing something else on another. The trade-off between exploitation versus exploration involves searching an area of the search space for the best solution or changing the area entirely. This concept is carried out often in other fields; therefore, research has already been done in some of these. 

A clear example of research comes from the domain of reinforcement learning: An agent needs to maximize the rewards obtained by selecting the actions that generate the most rewards in the long run. The best actions are initially unknown and must be discovered through trial and error. In the process of learning, the agent must select actions that seem to yield the best rewards (exploitation) while at the same time searching for better actions (exploration) for future selections. Neither exploitation or exploration, used independently of each other, will succeed at the task. Stochastic systems, moreover, require multiple observations of rewards from the same action to determine the underlying distribution governing the emission of rewards.

Within the area of reinforcement learning, multiple techniques have been developed with the intention of addressing the trade-off problem \cite{Sutton1998} along with some other problems at the same time\footnote{Other requirements might be to consider underlying distributions that vary over time for the emission of rewards, different available actions for different states, and receiving delayed rewards, among others.}. The problem of exploration versus exploitation in the reinforcement learning literature translates to selecting either the action with the highest estimated reward or some other available action. We proceed to briefly describe three families of methods for balancing exploration and exploitation.

\textit{Action-Value Methods} maintain an estimate of the ``goodness'' of an action by averaging the rewards obtained after selecting that action. The action that gets selected follows a greedy or $\epsilon$-greedy policy. A greedy policy is a set of rules that simply selects the action with the highest estimated value among all possible actions. $\epsilon$-greedy policies select the best action with a probability of $1 - \epsilon$ and any other random action with probability $\epsilon$. This ensures that all actions are selected and evaluated on the long run.

\textit{Reinforcement comparison methods} use a reference reward to compare all other rewards to estimate if a ``good'' or ``bad'' reward was obtained after following a particular action. Actions with bigger rewards are compared against actions with smaller rewards. The probability of selecting an action gets updates in relation to this comparison. As such, observing high rewards increases the probability of reselecting an action.

\textit{Pursuit methods} try to find the optimal action by making the greedy action more probable each iteration. This allows the true optimal action, given that a suboptimal action was thought to be the best, to catch up to the current best action.

\section{Self-organized Criticality}

The idea of criticality and self-organization was introduced in \cite{Bak1988}. Since then, self-organized criticality (SOC) has become a common approach for describing complex dynamic systems. Many domains of nature and social science have found that SOC models are able to represent a multitude of systems with varying degree of accuracy. Social sciences such as biology, astronomy, geography and economics have been able to model granular materials, evolution, earthquakes, landscape formation, traffic jams, brain functions, forest fires or solar flares using these type of systems \cite{Frigg2003}. SOC is considered by some \cite{Mora2011} a general theory\footnote{There is heated discussion in the scientific community regarding the validity of this fact \cite{Frigg2003} \cite{Wagenmakers2005}.} just as thermodynamics and Newtonian mechanics, for complex behaviours. But regardless of it not being a general theory, it still hold great value as modelling tools due to the fact that even flawed models can drive 
useful research and increase knowledge \cite{Frigg2003}.

Self-organized critical systems are those which are able to organize themselves naturally into a critical state without the aid of external sources. A critical state is a state where minor and locally constrained events are able to trigger the rearrangement of any number of elements in the system \cite{Bak1991}. According to this definition, the events triggering small rearrangement are the same triggering big and major rearrangement throughout the entire system. For instance, the movement distance of tectonic plates is not what defines the intensity of an earthquake; both small and catastrophic events are the results of the same locally constrained tectonic movement.

In section \ref{sec:soc_intro} we have mentioned that critical systems lie in the border of stability and instability. This also means that it is possible to observe two completely different behaviours depending on which side of the border we are in. When the system immediately converges to a stable state, we say the system is in a sub-critical configuration. On the other hand, when the system goes immediately into instability or unpredictability, the system is said to be in a super-critical region. We have mentioned before that SOC systems invariably exhibit a power-law distribution explaining some of its dynamics; notwithstanding, the existence of power-law distributions do not guarantee criticality. Beggs \etal propose one further test to prove the existence of criticality. If it is possible to find a specific set of parameters which would make the system sub-critical and another set that would make it super-critical, while at the same time observing a power-law distribution, then with high confidence we 
are dealing with 
a critical system \cite{Beggs2012}. This fact is important as it is used in section \ref{sec:adaptive_pso_performance} to prove the existence of criticality in a modified particle swarm.

A large collection of models explaining SOC exist throughout the literature. Most of the models, however, fall under two specific categories: the \textit{stochastic models} and the \textit{extremal models} \cite{Frigg2003}. In the next two sections these categories are further explained and illustrated with examples. Taking into account both models is important in this research as both are viable strategies for conferring particle swarms with self-organized criticality. On the one hand, if the topology of the swarm is treated as a dynamic system, the stochastic models are a good fit. On the other hand, if the particles are thought to be sharing information between each other through links of communication, the extremal models are a better fit.

\subsection{The Stochastic Models}

Stochastic models consist of stochastic dynamics operating with deterministic rules. A typical example of these is observed in the construction and dynamics of sand piles. Let us consider constructing a sand pile from scratch by placing individual sand particles on top of the pile at random locations (this accounts for the stochastic dynamics of the system). As the hight of the sand pile increases, newly added sand particles trigger the rearrangement of neighbouring local particles. If no sand is added, the pile remains static in a stable state (this accounts for the deterministic rules of the system). As more sand is added, a combination of growth and local particle rearrangements are observed. Suddenly, the addition of new particles affect not only local particles, but the sand pile globally. Chain effects are seen throughout the pile which involve small avalanches in different localities. The sand pile will stop growing. Every new particle added makes a particle slide. This is the self-organized critical 
state. The state is critical because any disturbance in the system (adding a new sand particle) makes the whole system unstable. It is self-organized because once the state has transitioned into a fully unstable state, without the intervention of external agents, the system can return to the critical state on its own.

\begin{figure} [ht]
\centering
\fbox {%
\includegraphics[scale=1]{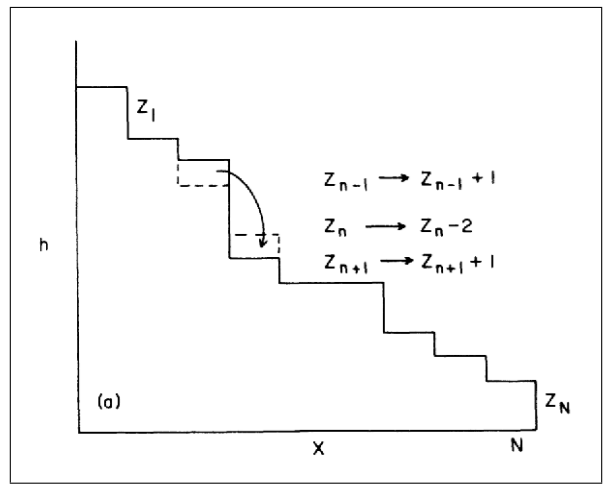}%
}
\caption{Dynamics of a Sand Pile.}
\label{fig:sand_pile_dynamics}
\end{figure}

The dynamics of sand pile models are simple in nature. Lets consider the illustration in Figure \ref{fig:sand_pile_dynamics}, taken from \cite{Bak1988}. This is an example of SOC in one dimension of size $N$. The system is bounded by a wall to the left, and sand can escape the system through the right. The number $z_n$ can be considered as height differences between two successive spatial steps along the abscissa: $z_n = h(n) - h(n+1)$, where $n = {1,2,...,N}$ and $h(i)$ is the height at location $i$. If sand is added in the $n^{th}$ position, we let
\begin{align*}
z_n & \rightarrow z_n + 1 \\
z_{n-1} & \rightarrow z_{n-1} -1.
\end{align*}
At any point in time, if the difference between heights ($z_n$) exceeds a constant maximum $z_c$, sand is rearranged such that
\begin{align*}
z_n & \rightarrow z_n - 2 \\
z_{n \pm 1} & \rightarrow z_{n \pm 1} + 1 \quad for \quad z_n > z_c.
\end{align*}
The critical point of this system is found when the slope formed by the sand pile is maximum, that is $\forall n \; z_n = z_c$. At this stage, if a particle is added it would tumble out of the system.

\subsection{The Extremal Models}
\label{sec:extremal_models}

Extremal models consist of deterministic dynamics operating with seemingly random rules. The most representative model of this category is the \textit{Bak-Sneppen} model of evolution \cite{Bak1993}. This model tries to explain the irregular mass extinctions observed in the apparently smooth process of evolution.

Most Darwinians believe that evolution is a continuous, smooth and gradual process \cite{Frigg2003}: the natural mechanisms of mutation and selection operate uniformly all the time. Detractors of the theory argue that the model is flawed as it cannot explain the known massive extinctions from the past. To explain these outlying events in the smooth process of evolution, external sources, such as extraterrestrial objects hitting the earth or massive volcanic eruptions, are cited. Jack Sepkoski found evidence suggesting that large extinctions follow a pattern \cite{Bak1993}. He later discovered the first clue of SOC in a system. If a histogram of the size of the extinctions is plotted against the frequency of these, a power-law distribution is observed. Extinction behaves, analogously, like avalanches from the sand pile model described in the last section.

As previously mentioned, the existence of a power-law does not guarantee SOC; therefore, a series of mathematical models were created to explain the observations. The most important of these models is the \textit{Bak-Sneppen} model. The set-up of the model is as follows. Consider a set $S$ of species. For the sake of simplicity, assume the species lie in a circular loop. Each specie is only able to interact with two neighbours. Species are assigned a fitness value in the interval $[0,1]$, where $1$ is the highest fitness available and $0$ the lowest. Each (discrete) time step the specie with the lowest fitness is extinguished (removed from the set) and replaced by another specie with a random fitness. Because the extinction of a specie will affect those neighbours interacting with it, the fitness of the neighbours is also replaced randomly.

The evolution of fitness in each specie and the number of chained continuous extinctions can be traced throughout multiple iterations. These traces show that during large periods of time things barely change. On the other hand, there are few periods where huge avalanches of extinction occur. The number of extinct species and the frequency of these extinctions follow a power-law distribution corresponding to the findings of Jack Seposki \cite{Frigg2003}, which suggest that evolution behaves as a SOC system. 

\subsection{The Power-Law Distribution}
\label{sec:powe_law}

We have previously made evident the importance of power-law distributions and SOC. In further chapters we try to confirm the existence of SOC in particle swarms by finding power-law distributions, among other things. For these reasons, understanding this distribution is crucial to this project. This section explains in detail this distribution.

Power-law distributions have been a major topic of interest over the years because of its mathematical properties. Many scientific observations using power-laws have been made describing complex systems such as earthquakes, the size of cities, the size of power outages, forest fires and the intensity of solar flares among many other natural and man-made phenomena \cite{Clauset2007}. Formally, a power-law distribution is observed when a random variable $X$ has the probability distribution
\begin{align}
\label{eq:power-law}
 p(X = x) = Cx^{-\alpha}.
\end{align}
The constant parameter $\alpha$ is known as the exponent or scaling parameter, satisfying the condition $\alpha > 1$; and $C$ is a normalization constant. It is often convenient to use a lower bound for which the law holds as the distribution diverges at zero. If we let the lower bound be $x_{min}$, the probability distribution, explicitly including the normalization constant, becomes
\begin{align}
 p(X = x) = \frac{\alpha - 1}{x_{min}} \left(\frac{x}{x_{min}}\right)^{-\alpha}.
\end{align}
This case holds for continuous values only. For discrete values the power-law distribution, stating again the normalization constant explicitly, is
\begin{align}
 p(X=x) = \frac{x^{-\alpha}}{\zeta(\alpha, x_{min})}, \\
 \zeta(\alpha, x_{min}) = \sum_{n=0}^{\infty}\left( n + x_{min} \right)^{-\alpha}.
\end{align}

\begin{figure} [ht]
\centering
\fbox{
  \subfloat[Using axes with a linear scale]{\label{fig:powerlaw_1}\includegraphics[scale=0.70]{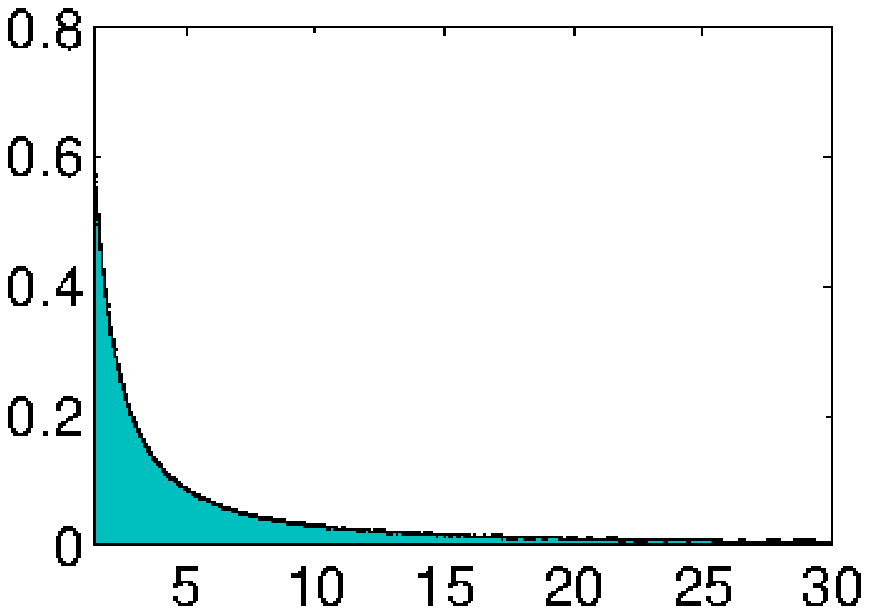}}
  \hspace{8mm}
  \subfloat[Using axes with a logarithmic scale]{\label{fig:powerlaw_2}\includegraphics[scale=0.70]{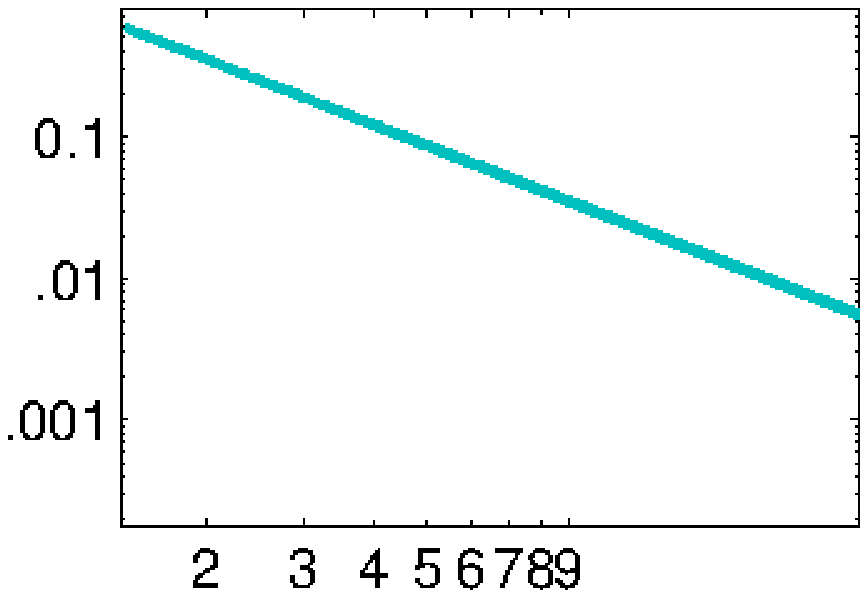}}
}
\caption{Graph of a Power-law distribution.}
\label{fig:power-law}
\end{figure}

Figure \ref{fig:power-law} shows how typical power-laws look in two different axis configurations. We have created 10,000 random data points from equation
\begin{align}
 f(x) = x^{\left(-\frac{3}{2}\right)}, \; x \in [0, 1].
\end{align}
Figure \ref{fig:powerlaw_1} shows the data points plotted against a linear scale in the ordinate and abscissa. If we take the logarithm of equation \ref{eq:power-law}, we obtain
\begin{align}
 \log P(X=x) = \log(C) - \alpha \log(x).
\end{align}
This is a straight line with a negative slope. Figure \ref{fig:powerlaw_2} shows the same data points plotted against a logarithmic scale in both ordinate and abscissa; as expected, it is a straight line.

There are multiple ways to approximate distributions; but specifically for power-law distribution, maximum likelihood estimate (MLE) seems to be the best option available. According to Clauset \etal. in \cite{Clauset2007}, all other approximations tend to end up being biased estimates for the parameters of power-laws. MLE, however, gives unbiased estimates in the long run as the size of the samples increase \cite{Clauset2007}. Throughout Chapter \ref{ch:evaluation}, power-laws are estimated with the tools provided by Clauset \etal for estimating power-law distributions in natural and man-made phenomena. In particular, the estimated parameters are $\alpha$ and $x_{min}$.

\section{Criticality and Particle Swarms}

In section \ref{sec:objectives} the objectives of this research were presented. These objectives revolve around particle swarms having self-organized criticality, but fair questions to ask are: Why would self-organized criticality be of any use to particle swarms or, generally, to heuristics? Are we expecting to improve particle swarms with criticality or just modify their behaviour? The answer to the second question is straightforward. The goal of conferring SOC to particle swarms is not to create a better variant, but rather, to observe its behaviour. A better variant, of course, would be a nice side effect; however, because the particle swarm is expected to modify its own parameters while searching for solutions, it could actually take longer than the standard particle swarm to achieve good results. The standard PSO version stagnates as soon as the particles get stuck inside local optimum \cite{Poli2009}. It is expected, therefore, for the SOC variant to perform better in the long run than the standard 
variant as SOC would enable large jumps to occur even after stagnation.

The first question has an empirical rather than an analytical answer. Criticality has been observed in multiple phenomena of nature. As such, many evolutionary heuristic algorithms mimicking nature have already incorporated this theory with good results \cite{Lovbjerg2002}. Genetic algorithms coupled with the concept of mass extinction are good examples of successfully applying SOC. Inspired by the model described in section \ref{sec:extremal_models}, SOC can be used to control the number of mutated genes in individuals and the level of extinction in the population. According to Krink \etal, the approach of applying SOC succeeded because continuous exploration is combined with focused exploitation, avoiding stagnation and loss of variance towards the final iterations \cite{Krink2000}.

  \chapter{Related Work}
\label{ch:related_work}

Research in the past has been done related to developing general parameter selection guidelines and incorporating parameter selection into particle swarm optimization algorithms (PSO). The most common approach to perform automatic parameter selection is to use meta-heuristic algorithms. This approach, instead of reducing the number of parameters, adds several more parameters in the form of hyper-parameters. These additional parameters are usually selected by hand according to parameter selection guidelines. For this reason this project is not concerned about this background material.

Other paths of research have explored the usage of critical limits (of chaos and stability) and the concepts of self-organized criticality (SOC) to create new variations of particle swarms. This chapter focuses mainly on exposing the research related to these specific topics. Proper parameter selection is also important and briefly discussed as it brings a solid base of understanding to the selection criteria of default parameter values.

\section{Proper Parameter Selection}

Before dealing with criticality and different methods for improving the exploratory and exploitive capabilities of the algorithm, we can take into account mathematical analysis for the selection of suitable parameters.

Plenty of research has been devoted to analysing the behaviour of PSOs given different sets of parameters. The analysis performed by Trelea in \cite{Trelea2003} is of high importance given that lower and upper bounds are derived for all parameters using only few assumptions. Trelea explores the impact parameters have in the performance of PSOs for different types of problems. Selection guidelines for the parameters are derived from his experiments. The convergence and divergence analysis of the paper are also of great importance. It explores which combination of parameters is likely to make the PSO diverge or converge. The upper and lower bounds for the parameters are used throughout chapter \ref{ch:design} for the design of a SOC PSO and are used as the default values in the program explained in chapter \ref{ch:implementation}. Table \ref{table:recommended_paramters} shows the recommended default parameter values given the balance they bring to the issues of exploration and exploitation.

\begin{table}[ht]
\centering    
\begin{tabular}{c | c}
Parameter & Default Value  \\ \hline
$\alpha_1$ & 1.494 \\
$\alpha_2$ & 1.494 \\
$\omega$ & 0.729  \\
$\Omega_{top}$ & 0.8  \\
$\Omega_{bottom}$ & 0.4 \\
\end{tabular}
\caption{Recommended parameter values for balancing exploitation and exploration.}
\label{table:recommended_paramters}
\end{table}

\section{Variants of the Particle Swarm Optimization Algorithm}

SOC has been growing in popularity since its inception in 1988 by the work of Bak \etal \cite{Kennedy1995}. From there, SOC has been growing in acceptance \cite{Poli2008}. It was natural for this concept to get incorporated into PSOs given the attention the scientific community has put into heuristics and SOC in the near past. 
Due to the scope of this project, we are only concerned in variants that relate to either the trade-off of exploration versus exploitation, or the concepts of criticality.

\subsection{Critical PSO}
\label{sec:critical_pso}

Lovbjerg and Krink \cite{Lovbjerg2002} devised a mechanism to incorporate one of the main concepts behind SOC: controlled divergence (or chaos). In this modification of the heuristic each particle has a criticality level $\mathcal{C}$ associated to it. If particles are close to each other, the criticality value increases; otherwise, it decreases with each iteration. Whenever the criticality of a specific particle reaches a maximum level, this criticality is dispersed to nearby particles and the original particle is randomly reallocated (or ``teletransported''). Particles relocated as a consequence of high critical values reset their criticality to the minimum value.

Furthermore, the criticality value is used to modify the exploratory capabilities of individual particles by modifying their inertia (the $\omega$ parameter in equation \ref{eq:pso_vel}). Instead of using a constant or linearly decreasing value (as some variants use), 
\begin{align}
\omega \rightarrow 0.2 + \frac{\mathcal{C}}{10}.
\end{align}
The idea behind this modification relates to the idea of close particles having less exploratory capabilities being, therefore, less diverse. The results of this modification are promising. The convergence capabilities are improved by adding a state of unpredictability to the particles (adding chaos). However, the relocation process of the particles and choosing the base value for $\omega$ add two new parameters which the user must tune.

\subsection{Chaotic PSO}


Building on top of the idea of adding chaos to a system to improve its performance, Liu et. al. in \cite{Liu2005} modified the particle swarm heuristic with the concepts of chaos. Chaos is a characteristic of non-linear systems where stochasticity emerges under deterministic conditions. Infinite unstable periodic dynamics are observed as a consequence. This particular variation of the particle swarm is named ``Chaotic PSO'' (CPSO). 

In the standard PSO, $\omega$ is the key to exploration and exploitation \cite{Liu2005}. With this idea in mind, chaos is applied to this parameter to make the system explore and exploit in a chaotic way. Adaptive inertia weight factors (AIWF) were introduced as a mechanism to explore or exploit depending on the relative fitness of each particle. If a particle evaluates well under an objective function (relative to all other particles), the inertia is lowered to allow exploitation. If, on the other hand, the objective value of a particle is below the average objective value of all particles, the inertia is increased to allow exploration. More rigorously,
\begin{align}
\omega \rightarrow & \omega_{min} + \frac{(\omega_{max} - \omega_{min})( f_{} - f_{min} )}{f_{avg - f_{min}}}, \; \text{for} \; f \leq f_{avg} \\
\omega \rightarrow & \omega_{max}, \; \text{for} \; f > f_{avg},
\end{align}
where $f$ is the function used to evaluate the fitness of a particle, $f_{avg}$ and $f_{min}$ are the average and minimum fitness of the entire swarm, respectively; and $\omega_{max}$ and $\omega_{min}$ are user specified parameters for the maximum and minimum values $\omega$ can take.

AIWF is used to allow exploration; chaos is added to the local search (CLS). The particles with the best objective value are duplicated and modified according to
\begin{align}
x_{n+1} \rightarrow \mu \times x_n (1 - x_n), \quad 0 \leq x_0 \leq 1, \label{eq:chaos_aiwf}
\end{align}
where $\mu$ is a control parameter and $n = {1,2,...,D}$. Equation (\ref{eq:chaos_aiwf}) is the implementation of a logistic function where chaotic dynamics are observed if $\mu = 4$ and $x \notin \{0.00, 0.25, 0.50, 0.75,1.00\}$.

\subsection{Evolutionary PSO}

In an attempt to remove the task of parameter selection, Miranda and Fonseca devised a way to adapt parameter selection using evolutionary algorithms in \cite{Liu2005}. The proposed heuristic is a modified version of the standard PSO that incorporates evolutionary operators to select parameters. Each evolutionary operator modifies the parameters of individual particles. Mutations modify parameter values randomly, replication copies a particle without any modification, and reproduction generates an offspring my mixing parameters according to the movement of the parents. The new algorithm finds good parameters for individual particles instead of the entire swarm.

  \chapter{Designing a Critical Particle Swarm}
\label{ch:design}

The main objective of this chapter is to present two new particle swarm optimization variants created with the intention to either observe criticality or an interesting behaviour in their dynamics. An interesting behaviour is that which balances in some way or form exploitation and exploration of solutions, and the ability of the swarm to keep exploring in the long run even under stagnation. In this project two PSO variants have been created. We begin by performing a dynamic system analysis on the PSO algorithm. The concepts and results of this analysis are what led to the creation of the first variant: the Eigencritical PSO. After having observed and analysed the behaviour of the first variant, a second variant was created using the conclusions drawn from evaluating the first variant. The second variant is named Adaptive PSO.

\section{Dynamic Systems Analysis}

Particle swarm optimization (PSO), as defined by equations
\begin{align}
\vec{V}_i(t+1) &{}= \omega \vec{V}_i(t)  + \alpha_1 \vec{r_1} \odot (\vec{P_i} - \vec{X_i}(t)) + \alpha_2 \vec{r_2} \odot (\vec{G} - \vec{X_i}(t)) \label{eq:pso_vel_2}  \\
\vec{X}_i(t+1) &{}= \vec{X}_i(t) + \vec{V}_i(t+1), \label{eq:pso_pos_2}
\end{align}
can be interpreted as stochastic dynamic systems. Each iteration particles move with stochastic dynamics around a ``fitness landscape'' searching for the best available fitness. The environment is completely unknown, the only available information is that which the particles have personally seen. Thus, the dynamics of the swarm heavily depend on the parameters $\alpha_1$, $\alpha_2$ and $\omega$ for a given configuration of $\vec{G}$ and $\vec{P_i}$. This can be more clearly observed by performing a dynamic system analysis on equation (\ref{eq:pso_vel_2}) and (\ref{eq:pso_pos_2}).

\subsection{Particle Swarms as Dynamic Systems}

Dynamic system theory states that the behaviour of the system depends on the eigenvalues of the transformation matrix used between successive iterations \cite{Trelea2003}. To be able to apply this theory, the normal PSO equations need to be reinterpreted with matrix operations. A few assumptions must be made in order to obtain a suitable set of equations translatable into matrix form. 

Instead of focusing in all dimensions at once, only one dimension will be considered at any single time. This assumption does not have any real impact on the algorithm as it already handles every dimension separately of each other, as can be appreciated by the usage of the component-wise multiplication operator $\odot$ in equation (\ref{eq:pso_vel_2}). For this analysis a deterministic set of equations is required, for this reason all random values are approximated with their expectation. PSOs only use random numbers generated from the same uniform distribution between one and zero; therefore, every random number is replaced by $\frac{1}{2}$. This is the second and last assumption made.

With these assumptions in mind, the following representation of the PSO is obtained:
\begin{align}
v(t+1) &= \omega v(t) + \frac{\alpha_1}{2}(p-x(t)) + \frac{\alpha_2}{2}(g-x(t)) \\
&= \omega v(t) + \frac{\alpha_1}{2}p - \frac{\alpha_1}{2}x(t) + \frac{\alpha_2}{2}g - \frac{\alpha_2}{2}x(t) \nonumber \\
&= \omega v(t) + \frac{(\alpha_1 + \alpha_2)\alpha_1}{(\alpha_1 + \alpha_2)2}p + \frac{(\alpha_1 + \alpha_2)\alpha_2}{(\alpha_1 + \alpha_2)2}g - \frac{(\alpha_1 + \alpha_2)}{2}x(t) \nonumber \\
&= \omega v(t) + \frac{(\alpha_1 + \alpha_2)}{2}\left( \frac{\alpha_1}{(\alpha_1 + \alpha_2)}p + \frac{\alpha_2}{(\alpha_1 + \alpha_2)}g - x(t) \right). \label{eq:dynamic_1}
\end{align}
For convenience the following is defined:
\begin{align}
\theta &= \frac{\alpha_1 + \alpha_2}{2}, \label{eq:dynamic_2} \\
\psi &= \frac{\alpha_1}{(\alpha_1 + \alpha_2)}p + \frac{\alpha_2}{(\alpha_1 + \alpha_2)}g. \label{eq:dynamic_3}
\end{align}
Combining equations (\ref{eq:dynamic_1}), (\ref{eq:dynamic_2}) and (\ref{eq:dynamic_3}) yield the final pair of equations:
\begin{align}
v(t+1) = \omega v(t) + \theta(\psi - x(t)), \\
x(t+1) = x(t) + v(t+1).
\end{align}
Notice how the variables $v$, $x$, $g$ and $p$ are now scalars (represented with lower-case letters to avoid confusion), in contrast with equations (\ref{eq:pso_vel_2}) and (\ref{eq:pso_pos_2}) where they are vectors.

Equations (\ref{eq:pso_vel_2}) and (\ref{eq:pso_pos_2}) are ready to be reinterpreted in matrix form as follows:
\begin{align}
&\vec{y}(t + 1) = \mathcal{A}\vec{y}(t) + \mathcal{B}\psi \text{, having} \\
&\vec{y}(t) = \left[ \begin{array}{c}
            x(t) \\
            v(t)
           \end{array}  \right], \;
\mathcal{A} = \left[ \begin{array}{cc}
                      1 - \theta & \omega\\
                      -\theta & \omega
                     \end{array} \right], \;
\mathcal{B} = \left[ \begin{array}{c}
                     \theta \\
                     \theta
                     \end{array}  \right].
\end{align}
In the context of dynamic systems analysis, the vector $\vec{y}$ is referred to as the \textit{state} of a particle, $\mathcal{A}$ is the \textit{dynamic matrix}, $\psi$ is the \textit{external input}, and $\mathcal{B}$ is the \textit{input matrix} \cite{Trelea2003}. Matrix $\mathcal{A}$ is responsible for the converging or diverging dynamics of the system throughout two consecutive iterations. $\psi$ drives the particles towards a specific point. The matrix $\mathcal{B}$ applies the influence of the external input to the particles.

A stable system is that which has all eigenvalues of matrix $\mathcal{A}$ less than one. In contrast, if one or more eigenvalues are above one, the system is divergent. If the eigenvalues are imaginary, the system tends to oscillate into stability or to diverge, depending on the real part. In other terms (more closely related to criticality and our topic of interest), if all eigenvalues are below one, the system is stagnant; otherwise, the system is chaotic. This is the first clue for conferring criticality to a PSO.

According to this analysis, it should be possible to obtain interesting behaviours out of a PSO if the highest eigenvalue of the dynamic matrix $\mathcal{A}$ is one. This should account for a system which is neither completely stagnant nor entirely chaotic. A variant of the standard PSO, named Eigencritical PSO, has been created with these properties in mind. The following section describes this algorithm and some of its characteristics.

\section{Eigencritical Particle Swarm}

In the previous section it was shown how the eigenvalues of matrix $\mathcal{A}$ could be the key for making a PSO critical. This PSO variation has been created to test this hypothesis. 

If the assumptions made in the previous section are considered (using the expected value of random numbers and calculating each dimension of the particles independently of each other), the algorithm trades its stochasticity for the capability to have a fixed and constant set of parameters that make the eigenvalues of matrix $\mathcal{A}$ one. Loosing the stochastic behaviour is a bad idea \cite{Spears2010}; this would not only make the algorithm boring, but useless too. The algorithm needs to remain stochastic and, somehow, adapt the parameters according to the stochastic behaviour to achieve the desired dynamic matrix $\mathcal{A}$.

Instead of trying to adapt the parameters every iterations in such a way that would make the eigenvalues of matrix $\mathcal{A}$ equal to one for the next iteration, the highest eigenvalue is ``forced'' to always be one without tampering with parameter adaptation. The ``forcing'' procedure is as follows. The state $\vec{y_i}(t)$ of every particle $i$ for all dimensions at time $t$ is observed. The standard PSO algorithm is then executed using the default parameters specified in table \ref{table:recommended_paramters}. The new state of the particles $\vec{y_i}(t+1)$ for every dimension is recorded but the particles are not modified. The particles stay as they were at time $t$.

Using the present ($\vec{y_i}(t)$) and future ($\vec{y_i}(t+1)$) information of all particles and dimensions, the following equation is constructed:
\begin{align}
\label{eq:eigencritical_equation}
 \left[\begin{array}{c}
  \vec{X_1}(t+1) \\
  \vec{X_2}(t+1) \\
  \vdots \\
  \vec{X_N}(t+1)
 \end{array}\right]
 =
 \mathcal{C}
 \left[\begin{array}{c}
 \vec{X_1}(t) \\
 \vec{X_2}(t) \\
 \vdots \\
 \vec{X_N}(t)
 \end{array}\right],
\end{align}
where $X(t) = \{x^{(1)}(t), x^{(2)}(t), ..., x^{(D)}(t)\}$, $D$ is the number of dimensions of the particles and $N$ the total number of particles.

For every iteration of the Eigencritical PSO, equation (\ref{eq:eigencritical_equation}) is solved for $\mathcal{C}$ by an approximation method. Because the system has the form $y = Ax$, and solving for $A$ is required, the approximation is performed against $y^T = x^TA^T$. There are multiple methods which can be used to approximate $\mathcal{C}$. The approximation has to be performed every iteration, given this limitation, a householder method using QR decomposition is used as it is fast and accurate enough. The reader is referred to \cite{Bischof1989} for more information regarding this method.

The eigenvalues $\vec{\lambda}$ of matrix $\mathcal{C}$ are obtained and the highest eigenvalue $\lambda_1$ is used to modify the matrix as follows:
\begin{align}
\label{eq:eigencritical_C_modification}
\mathcal{C} \rightarrow \mathcal{C} \frac{1}{\lambda_1}.
\end{align}
This guarantees that the highest eigenvalue is always one because, by the definition of eigenvalues and eigenvectors, we have
\begin{align*}
 \frac{\lambda}{\lambda}v = \frac{\mathcal{C}}{\lambda} v,
\end{align*}
where $v$ is an eigenvector and $\lambda$ an eigenvalue.

After this step, the state of the particles is modified by applying the transformation $\mathcal{C}$ to the position of all particles. As such, the following operation is performed:
\begin{align}
\label{eq:eigencritical_C_application}
 \vec{X}(t+1) = \mathcal{C} \left[\vec{X}(t)\right].
\end{align}
This time, however, the position of the particles is modified and a new iteration starts again.

This is a summary of the algorithm performed by the Eigencritical PSO.
\begin{enumerate}
 \item Initialize the swarm just as the standard PSO is initialized.
 \item Run the standard PSO variant with default parameters. Observe the resulting (or future) position of the particles, but do not modify the present position of the particles.
 \item Construct equation (\ref{eq:eigencritical_equation}) using the present and future information of the particles.
 \item Solve for $\mathcal{C}$ with an approximation method as there might not be a possible solution.
 \item Find $\lambda_1$, the largest eigenvalue of $\mathcal{C}$.
 \item Modify $\mathcal{C}$ with equation (\ref{eq:eigencritical_C_modification}), to make the largest eigenvalue of the matrix equal to one.
 \item Apply the transformation $\mathcal{C}$, as described by equation (\ref{eq:eigencritical_C_application}), to the present position of the particles, modifying their state.
 \item Repeat from 2, until a desired number of iterations have been performed.
\end{enumerate}

\section{Adaptive Particle Swarm}
\label{sec:adaptive_pso}

In Chapter \ref{ch:evaluation}, the Eigencritical PSO variation is tested and some interesting behaviours are observed. The details of these tests and results are left for later; however, these observations motivated this particle swarm variant. It was observed that having the biggest eigenvalue equal to one modified the exploratory and exploitative behaviour of the swarm in the event of stagnation. The second crucial observation is that modifying parameters is all it takes to achieve an interesting critical behaviour; there is no need to further modify the PSO algorithm with new parameters. These are all desirable traits and some of the objectives behind this variation.

This variation, aptly named Adaptive PSO, tries to adapt the parameters of the algorithm on-the-fly with the objective of setting the largest eigenvalue of the observed transformation equal to one. The parameters are modified according to the dynamic behaviour of the swam. Instead of calculating the eigenvalues; nonetheless, a different metric is used to asses if the swarm is either converging or diverging. Calculating eigenvalues is an expensive operation which cannot be afforded every iteration if the PSO variant wants to be efficient.

Multiple metrics can be chosen to efficiently analyse the dynamics of the swarm. Three different metrics have been designed to work with this PSO version. Metric one measures the average distance between every particle of the swarm. Metric two measures the average distance between every particle and the centroid of the swarm. Finally, the third metric measures the average velocity norm of all particles. The first two metrics are concerned about the size of the swarm and the thirds is concerned about the speed at which the particles move. It is obvious how the first two metrics measure the dynamics of the swarm: if the difference of these metrics between two successive iterations is positive, the swarm is growing or diverging; if, on the other hand, the difference is negative, the swarm is collapsing or stagnating. The third metric measures the dynamics of the swarm in a slightly different way. The velocity of the particles is the core of the PSO algorithm, as it is evident from equation (\ref{eq:pso_vel_2}). 
By measuring the norm of the velocity, in essence, the ability of the particles to move around is being assessed. The difference of this metric between two successive iterations quantifies the exploratory and exploitive behaviour of the swarm. If the difference is positive, the swarm has a tendency to explore, if the difference is negative, the swarm tends to exploit.

For the purpose of normalizing all metrics between the interval $[-1, 1]$, the following sigmoidal function is used:
\begin{align}
&\eta = \frac{B}{2}, \nonumber \\
&f(x) = \left[ \frac{1}{\left(1 + exp(-\frac{x}{\eta}) \right) - 0.5} \right] \times 2, \label{eq:adaptive_sigmoid}
\end{align}
where $B$ is the boundary size of the search problem. The form of this function has been selected because of several desirable properties. The difference between metrics is not a bounded number; in theory this difference could be anywhere in between $[-\infty, \infty]$. The standard sigmoid function\footnote{The sigmoid function is defined as $\sigma(x) = \frac{1}{1 + exp(-x)}$.} ``squashes'' the supplied $x$ into the range $[0, 1]$, and we are interested in the range of $[-1, 1]$. $\eta$ is a stretching parameter used to adapt the sensibility of equation (\ref{eq:adaptive_sigmoid}) to large values. It is assumed that the particles do not need to explore beyond the boundary the PSO uses to randomly initialize the position of the particles in the first iteration; therefore, this parameter is automatically set using the boundary size. For this and other experiments performed, the boundary size describes the radius of a hypersphere around the origin where particles are allowed to be randomly initialized.

The metrics are used to modify the parameters; nevertheless, multiple rules can still be defined to perform this modification. The difference of metrics is formally defined as
\begin{align}
\Delta S &= S(t+1) - S(t), 
\end{align}
where $S(t)$ is the metric observed at time $t$, and $S(t+1)$ the metric at time $t+1$. Two rules are proposed to modify the parameters. The first rule is defined as
\begin{align}
 \theta \rightarrow \theta - (\varepsilon \times \Delta S), \; \varepsilon < 1.
\end{align}
Here, $\theta$ takes the place of any parameter, and $\varepsilon$ is a new parameter introduced to regulate the size of the modification. This rule modifies all parameters with the same amount each iteration. The second rule explores the alternative of modifying each parameter proportionally to its current value. This rule is defined as
\begin{align}
\theta \rightarrow \theta - \left[ (\varepsilon \times \Delta S) \times \theta \right], \; \varepsilon < 1.
\end{align}
Yet again, $\theta$ represents any parameter and $\varepsilon$ works as a constriction factor for the size of the parameter modification. The new parameter $\varepsilon$ is referred to as the \textit{adaptive epsilon}, rule one is known as the \textit{dependant rule}, and rule three as the \textit{independent rule}.

In both rules, if $\Delta S$ is positive; meaning that the swarm is diverging, the parameters are reduced in size to try and avoid chaos. If $\Delta S$, on the other hand, is negative, the swarm is collapsing and the parameters are modified positively to avoid stagnation.

After the parameters have been modified, the PSO is executed normally. What follows is a summary of the Adaptive PSO variant described in this section.
\begin{enumerate}
 \item Initialize the swarm just as the standard PSO is initialized.
 \item Set the metric $S$, among the three available, to use throughout the run.
 \item Set the rule $R$ to modify the parameters.
 \item Set the new parameter $\varepsilon$ according to a user supplied value.
 \item Initialize the parameters $\vec{\theta} = \{ \alpha_1, \alpha_2, \omega \}$ with arbitrary default values.
 \item Run the standard PSO variant using the parameter set $\vec{\theta}$.
 \item Measure the dynamics of the swarm using the selected metric $S$.
 \item Modify the parameters $\vec{\theta}$ according to the user selected rule $R$.
 \item Repeat from 6, until a desired number of iterations have been performed.
\end{enumerate}
\chapter{The Experimentation Platform: PSO Laboratory}
\label{ch:implementation}

A test platform has been developed as part of this project to perform some general research on particle swarms and to test the two new variations created in the previous chapter. This chapter describes the characteristics of this program as well as its graphical user interface.

\section{The Program}

PSO Laboratory is a program entirely developed in C++ and the Qt graphics library with multiple objectives in mind. The program performs all operations in real-time and allows user interaction to control parameters on-the-fly. Wherever it was beneficial, multiple threads are used to perform operations faster and in an efficient manner. The program is optimized for performance and not necessarily for memory usage. Because it is hoped that the program is useful not only for this project, but for other researchers in the field of particle swarms, attention has been put to the graphical user interface, usability and programming details. The code is heavily commented and documented with a focus on reusability and extensibility.

\subsection{The Graphical User Interface}

\begin{figure}[ht!]
\centering
\fbox {%
\includegraphics[width=0.9\textwidth]{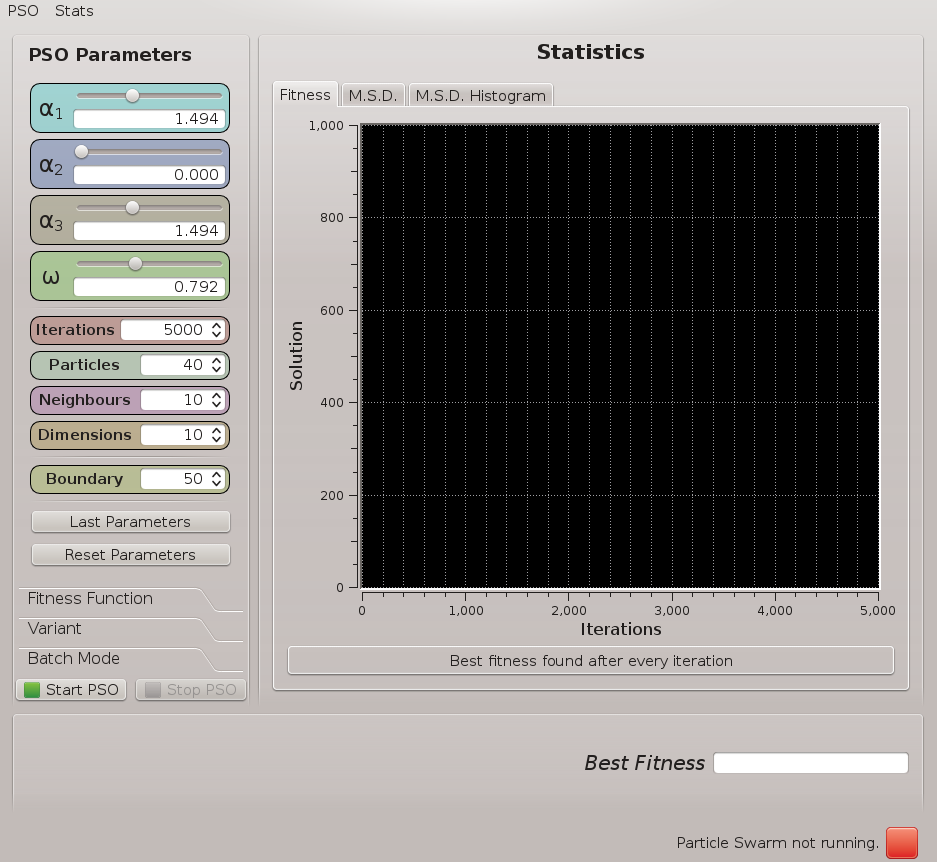}%
}
\caption{Main window of the PSO Laboratory.}
\label{fig:psolab_gui}
\end{figure}

The graphical user interface is divided in two main sections. Figure \ref{fig:psolab_gui} shows the main window of the program.
The left side of the window holds a series of tabs which contain parameters and options the user can select to modify the behaviour and characteristics of the desired PSO algorithm. The right side holds graphical tools useful for observing the real-time behaviour of the swarm. These graphical tools are explained in section \ref{sec:graphs_and_plots}. What follows is a description of each tab in the left hand side.

\subsubsection{PSO Parameters Tab} 
\label{sec:pso_parameters}

The parameters common to all particle swarms are found in this tab. These parameters are divided into three sections. The upper section contains the parameters $\alpha_1$, $\alpha_2$, $\alpha_3$ and $\omega$. In Chapter \ref{ch:background} and \ref{ch:design} only the parameters $\alpha_1$, $\alpha_2$ and $\omega$ were introduced. The parameter $\alpha_3$ has no relevance to this project and will always be set to zero. This parameter controls a component found in the neighbourhood PSO variant and was included in the program as part of the objective of making it more general and useful not only to this project, but to others too. We refer the reader to \cite{Suganthan1999} for more information regarding this variation and the usage of this parameter.

The parameters $\alpha_1$ and $\alpha_2$ are bounded by the interval $[0, 4]$, which, according to \cite{Trelea2003}, is considered a suitable range for these parameters. In the same way, $\omega$ is bounded by the interval $[0,2]$. When the standard PSO or the Eigencritical PSO is selected, these parameters can be modified on-the-fly while the PSO algorithm is running to immediately observe the impact the changes have on it. To modify the parameters the slider can be dragged or a number can be typed into the textbox (after which the enter key needs to be pressed) corresponding to the desired parameter value.

The middle section of this tab hold the parameters used to initialize all particle swarms. The \textit{iterations} parameter specifies the number of iterations the algorithm will perform. \textit{Particles} specifies the number of particles used in the swarm. \textit{Neighbours} is used only if the parameter $\alpha_3$ is non-zero and is not relevant to this project. It specifies the number of neighbours to use instead of the entire swarm to select the global best known location for each particle. \textit{Dimensions} indicates the number of dimensions to use for the functions that evaluate the fitness of the particles. This value indirectly specifies the difficulty of the functions: higher dimensions make the ``fitness landscape'' bigger, more sparse and more difficult to explore. Finally. \textit{Boundary} specifies the radius of a hypersphere with centre at the origin which specifies a bounded location where particles are able to be placed in their initial random placement.

The bottom section contains only two buttons. These are meant to help the researcher keep track of the initial parameters used to start a PSO. Because the parameters can be modified on-the-fly while a PSO is running, or adapted by the Adaptive PSO variation, the initial parameters are lost. Instead of having to specifying the original parameters again, the button \textit{Last Parameters} restores these to the values used in the last run. The button \textit{Reset Parameters} sets the parameters to the default values displayed when the program is first executed. This default values are the ones specified in table \ref{table:recommended_paramters}.

\subsubsection{Fitness Function Tab}

\begin{figure}[ht!]
\begin{center}
\begin{tabular}{m{4.5cm} m{5.5cm}}
\multicolumn{2}{l}{\textbf{Sphere Function}} \\
\includegraphics[scale=0.4]{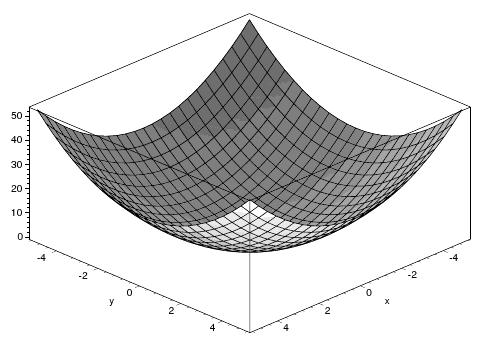} &$\displaystyle\sum\limits_{i=1}^{N} x_i^2$ \\
\end{tabular}
\begin{tabular}{m{4.5cm} m{5.5cm}}
\multicolumn{2}{l}{\textbf{Rastrigin Function}} \\
\includegraphics[scale=0.4]{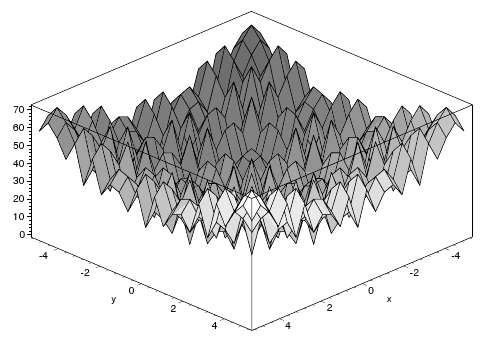} &$\displaystyle\sum\limits_{i=1}^{N}\left(x_i^2 - 10cos(2 \pi x_i) + 10 \right)$ \\
\end{tabular}
\begin{tabular}{m{4.5cm} m{5.5cm}}
\multicolumn{2}{l}{\textbf{Grewank Function}} \\
\includegraphics[scale=0.4]{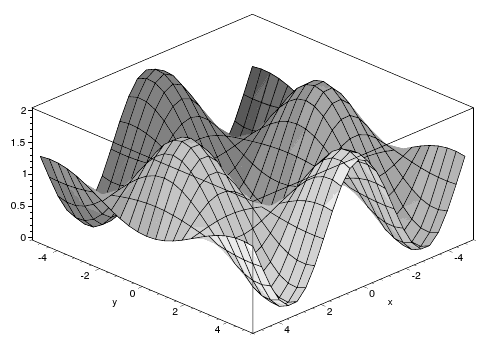} &$\frac{1}{4000} \displaystyle\sum\limits_{i=1}^{N}{x_i^2} - \prod_{i=1}^{N}{cos \left( \frac{x_i}{\sqrt{i}}\right)} + 1$ \\
\end{tabular}
\begin{tabular}{m{4.5cm} m{5.5cm}}
\multicolumn{2}{l}{\textbf{Schwefel Function}} \\
\includegraphics[scale=0.4]{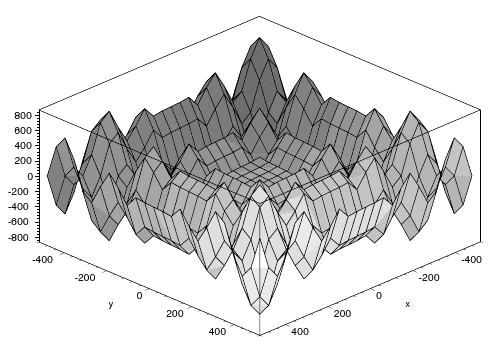} &$\displaystyle\sum\limits_{i=1}^{N} \left[ -x_i sin(\sqrt{ \left| x_i \right| }) \right] $ \\
\end{tabular}
\end{center}
\caption{Available fitness functions in PSO Laboratory.}
\label{fig:fitness_functions}
\end{figure}

The fitness function tab specifies four functions which can be used to test the PSO algorithms. Figure \ref{fig:fitness_functions} shows a two-dimensional representation of each available function along with its corresponding equation. In every equation, $N$ represents the total number of dimensions.

The sphere function is the simplest of all available functions, serving the purpose of being a simple test scenario and a comparison mechanism. It has a single global minimum at $0$, and is achieved when all dimensions are $0$. The Rastrigin function has frequent, evenly distributed, local minima. Its global minimum is found at $0$ when all dimensions are $0$. The Grewank function shares some similarities with the Rastrigin function. It is highly multimodal with local minima evenly distributed. Depending on the scale; however, the function has very different shapes. At large scales the function appears to be convex, as the scale is reduced, more and more local extremum are observed. The global minimum is found at $0$ when all dimensions are $0$. The Schwefel function has been selected as part of the test functions as it has, in contrast with the others, its global minimum at $-418.9829N$ when all dimensions are $-420.9687$. This is a deceptive function since the global minimum is far away from the next best 
local minimum and, furthermore, at the origin lies a relatively big flat valley. 

\subsubsection{Variant Tab}

This tab allows the user to select a PSO variant to test. There are three variants available: the standard PSO, the Eigencritical PSO and the Adaptive PSO. After selecting one of these variants, if there are tunable options relevant to that version only, more parameters are immediately displayed. The parameters available and their description are presented in later sections.

\subsubsection{Batch Mode Tab}

With the purpose of automatically testing multiple times the same PSO variant using a specific set of parameters, a batch mode has been developed. This mode allows the user to run the program with a predefined configuration multiple times without requiring user interaction. In this mode the real-time graphs are disabled and the results of all runs are stored in a user specified folder for later analysis. This mode uses as many threads as the host computer supports to speed up the process. Details of this mode are explained in a later section.

\section{The PSO Variants}

This section explains the option available to the user for every PSO variation.

\subsection{Standard and Eigencritical PSO}

Both standard and eigencritical PSO variations share the same number of parameters. These parameters were already described in section \ref{sec:pso_parameters}. There are, nonetheless, important considerations to take into account when dealing with the Eigencritical PSO variant. This variation uses the parameters in a different way than any other PSO. In every iteration these are used to calculate the velocity of the particles with equation (\ref{eq:pso_vel_2}), but this velocity is not used to calculate the next position of the particles. Instead, the velocity is used to calculate the future position of the particles for which a linear transformation from the present position to the future position is calculated. In the standard PSO the parameters directly modify the behaviour of the swarm, in the Eigencritical PSO, however, the parameters indirectly modify the behaviour by changing the inherent difficulty of finding a linear transformation.

Some parameter combinations have the desired effect of allowing a linear transformation with a low mean square error to be found; on the other hand, some set of parameters make the linear transformation inaccurate (with a high error value). When a linear transformation is not possible, because of the behaviour of the Household QR decomposition used to calculate the transformation, a least squares approximation is obtained.

\subsection{Adaptive PSO}

\begin{figure}[ht!]
\centering
\fbox {%
\includegraphics[scale=1]{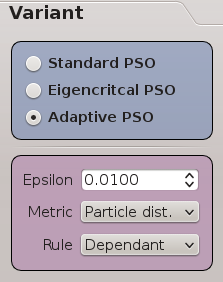}%
}
\caption{Parameters exclusive to the adaptive PSO variant.}
\label{fig:adaptive_pso_params}
\end{figure}

The Adaptive PSO variant uses all parameters described in section \ref{sec:pso_parameters} as well as some other exclusive ones. When this version is selected in the graphical user interface, a new box containing parameters is shown in the variants tab. Figure \ref{fig:adaptive_pso_params} shows the box with the new parameters. \textit{Epsilon} corresponds to the adaptive epsilon parameter, \textit{Metric} to the metric used for modifying the parameters, and \textit{Rule} for the rule used when modifying the parameters with the selected metric. These parameters were described in detail in section \ref{sec:adaptive_pso}.

The available metrics are \textit{Particle dist.}, which corresponds to the average distance between all particles, \textit{Centroid dist.}, corresponding to the average distance from every particle to the centroid of the swarm, and \textit{Vel. norm}, the average velocity norm of all particles in the swarm. 

The two rules available for modifying the parameters correspond to the \textit{Dependant} and \textit{Independent} rules described in section \ref{sec:adaptive_pso}.

One significant difference between this variant and the previous two is that the possibility of modifying the standard parameters on-the-fly is no longer possible. After this PSO starts running, the parameters are locked and no user interaction is possible. However, because this PSO modifies its own parameters, in the PSO Parameters tab it is possible to see how the parameters are being adapted.

\section{Tools for Data Analysis}

PSO Laboratory was designed with the objective of supplying a test platform to run PSO algorithms; the responsibility of analysing data quantitatively is left for other tools such as Matlab. Some basic tools, nevertheless, are provided to observe the real-time behaviour of the swarm in the form of graphs. All data generated by the program can be dumped into a CSV (comma separated values) for later analysis. This sections describes the graphical tools provided and the feature of dumping data into a CSV file.

\subsection{Graphs and Plots}
\label{sec:graphs_and_plots}

The available plots are updated in real-time while a PSO runs. To avoid slowing down the PSO, the plots are completely detached from the algorithm. The plots sample the PSO algorithm for information every 2 milliseconds. Because more than one iteration of the algorithm can be executed in between these 2 milliseconds, the information shown is an overview of the underlying behaviour. This method of sampling has two main advantages: one, the graphs are visually less cluttered and often times easier to interpret; and two, an effect of smoothing the received information is achieved.

The plots available are the fitness graph, mean square distance (MSD) graph and MSD histogram. In the first two plots, the mouse can be left-clicked and dragged to zoom into an area. Multiple zooms can be stacked. The middle-mouse button goes back one zoom level; the right-mouse button reset the zoom in its entirety.

The fitness graph shows the best solution found so far along the execution of the algorithms. Accompanying this graph, in the bottom section of the main window, a text box with the label ``Best Fitness'' displays the current best solution found so far. Every iteration the size of the swarm is measured by calculating the MSD. from every particle to the centroid of the swarm. The MSD graph plots the size of the swarm for every sampled iteration. This information is important for detecting criticality and observing how the swarm is exploring and exploiting. Furthermore, from this graph it is possible to tell if the particles are stagnating or diverging. The MSD histogram plots the positive increments of the swarm size between two consecutive samples against the frequency of these differences. There are two tunable parameters for this graph. \textit{Bin Size} allows the user to change the range used for grouping the size differences, and \textit{Log Scale} transforms the ordinate and abscissa from a linear 
scale to a log scale. This last parameter is specially useful for detecting criticality in the size dynamics of the swarm. As it was explained in section \ref{sec:powe_law}, in a log-log scale, a line with negative slope should be observed if a power-law distribution exists. These parameters, just as every other parameter, modify the graphs in real-time.

\subsection{Capturing and Dumping Run Statistics}

\begin{figure}[ht!]
\centering
\fbox {%
\includegraphics[scale=2]{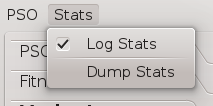}%
}
\caption{Options available under the Stats menu-item.}
\label{fig:pso_stats}
\end{figure}

In the menu-bar of the main window, as shown in Figure \ref{fig:pso_stats}, the \textit{Stats} menu-item is available. The \textit{Log Stats} checkable option enables or disables the recording of statistics throughout the execution of a PSO. If this option is disabled, while the PSO algorithm runs less memory is used, but the complete information related to the run is not stored (and therefore cannot be accessed later by external applications). If this option is checked, on the other hand, the entire information of the run is stored and by pressing the button \textit{Dump Stats} the information can be saved into a CSV file. 

\section{Batch Operations}

Batch operations provide the means for executing PSOs using a predefined set of parameters multiple times and simultaneously without user interaction. The batch operations tab holds all parameters and information needed to create a set of identical PSOs which can be run in parallel. The program automatically detects the available number of real kernel threads and uses all to run the PSOs. The executed PSOs pick up the parameters set by the user, just as if a single PSO was run in a standard way. This project uses this feature to generate enough data for every desired PSO configuration to obtain statistically significant results.

\begin{figure}[ht!]
\centering
\fbox {%
\includegraphics[scale=1]{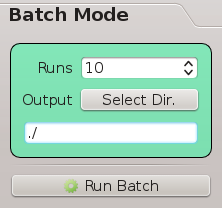}%
}
\caption{Options under the Batch Mode tab.}
\label{fig:batch_tab}
\end{figure}

Figure \ref{fig:batch_tab} shown the options available under the batch operations tab. The option \textit{Runs} specifies the number of times to execute a PSO using the predefined parameters set by the user. The \textit{Select Dir.} button, attached to the \textit{Output} label, displays a dialogue where the user can select a directory for storing the information generated by each individual PSO. By default the current directory (``./") is selected. For each PSO run, a CSV file with the format ``swarm\_xxx.csv'' is generated, where ``xxx'' is an integer padded (to the left) with as many zeros as required to fill the three available characters. Below this button, a text box displays the directory where the PSO information will be dumped.
  \chapter{Experimental Results and Discussion}
\label{ch:evaluation}

The theory described in previous chapters is put into practice here. The standard PSO is tested and compared against the results found in other works with the intention of verifying the correctness of the algorithm, as all other variants are based on this one. The behaviour of the Eigencritical PSO is examined in detail and tests are implemented to evaluate its performance compared to the standard particle swarm. The Adaptive PSO, with its multiple variations and modifications, is examined and evaluated against the standard and Eigencritical PSO. Four different problems are used to evaluate the performance and behaviour of all particle swarms while searching for traces of self-organized criticality in the dynamics.

\section{Testing the Standard PSO Implementation}

This section serves two purposes. On the one hand, the standard PSO is compared against results from other works; on the other hand, the different plots used to present results in the following sections are introduced and explained. The standard PSO algorithm is simple, yet crucial to the development of this work. It is important, therefore, to test the implementation for correctness. The methodology for comparing the standard PSO is as follows: Two works that test the standard PSO are chosen and the results are attempted to be replicated. For each test, the PSO parameters are set exactly as indicated by each independent test. Whenever a parameter is not specified, suitable values are searched for such as to mimic the expected results. For each test, the algorithm is executed 50 times for statistical purposes.

\subsection{First Standard PSO Test}

Lovbkerg \etal in the research \cite{Lovbjerg2002} experiment with PSOs by adding a new component to the algorithm which would behave in a critical manner. Section \ref{sec:critical_pso} explains this research in some more detail. In multiple instances the standard PSO is tested against newly created variations. Figure \ref{fig:comparison_rastrigin} shows the results obtained by the standard PSO and two other variants for the Rastrigin objective function using 30 dimensions. The result marked with ``STD Average Best'' is the only one relevant to this comparison. This result shows the average fitness over multiple runs of the standard PSO.

\begin{figure}[ht!]
\centering
\fbox {%
\includegraphics[scale=1.2]{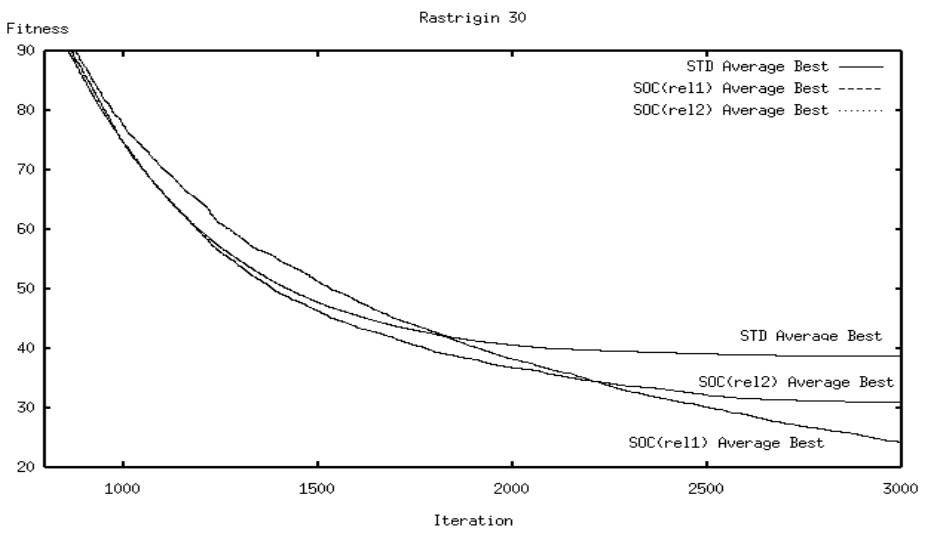}%
}
\thesiscaption[Standard PSO and the Rastrigin objective function.]{Results obtained by Lovbkerg \etal using the standard PSO with the Rastrigin objective function.}{fig:comparison_rastrigin}
\end{figure}

Figure \ref{fig:comparison_rastrigin_mine} shows detailed results of running the standard PSO 50 times with the PSO Laboratory program, including all plots relevant to this project. 

\begin{figure}[ht!]
\centering
\fbox {%
\begin{tabular}{cc}
\subfloat[Mean fitness of PSOs.]{\label{fig:comp_1_mean_fitness}\includegraphics[scale=0.70]{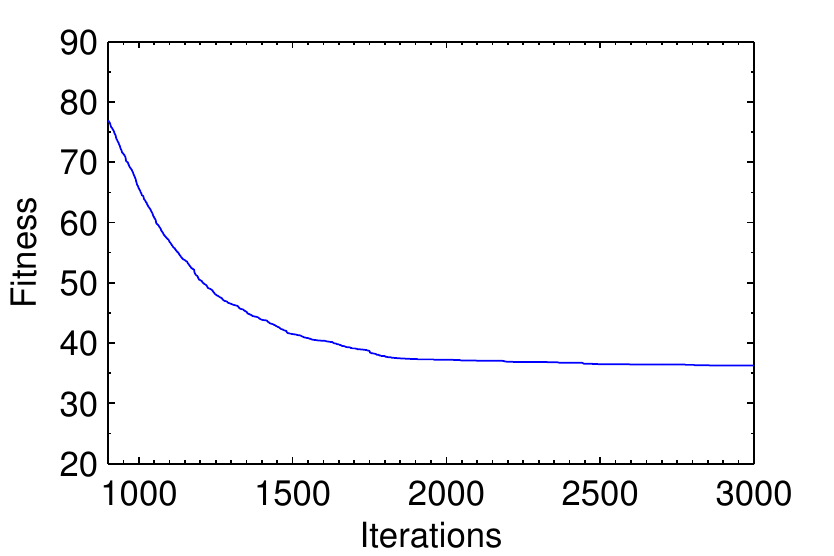}}%
&
\subfloat[Individual fitness of all PSOs.]{\label{fig:comp_1_all_fitness}\includegraphics[scale=0.70]{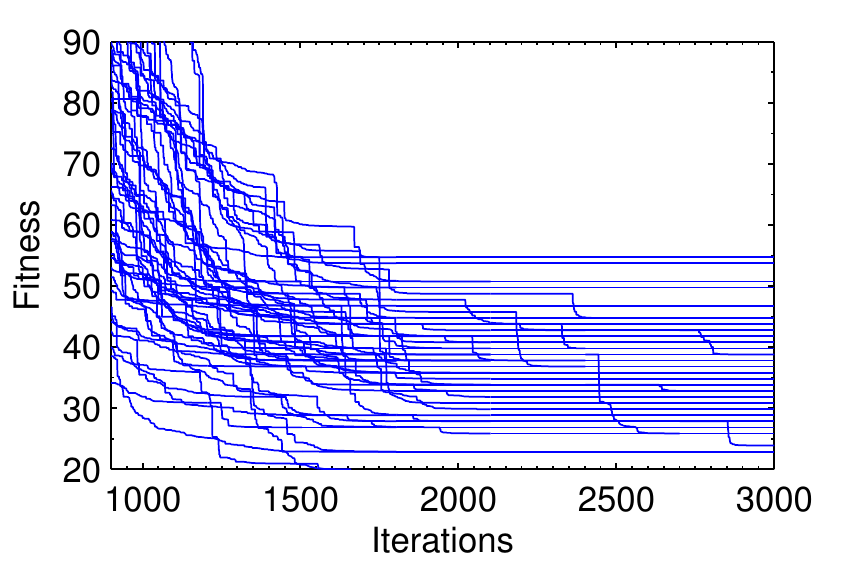}}\\
\subfloat[Mean MSD of all runs.]{\label{fig:comp_1_mean_msd}\includegraphics[scale=0.70]{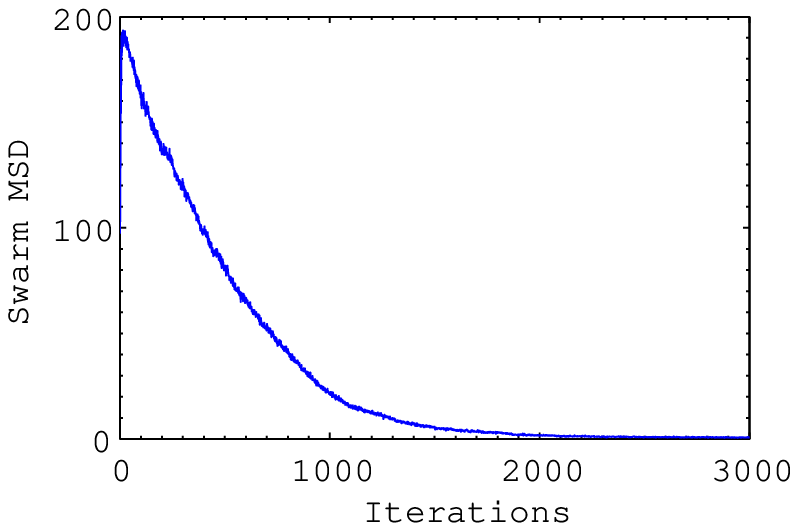}}%
&
\subfloat[Power-law approximation.]{\label{fig:comp_1_power-law}\includegraphics[scale=0.70]{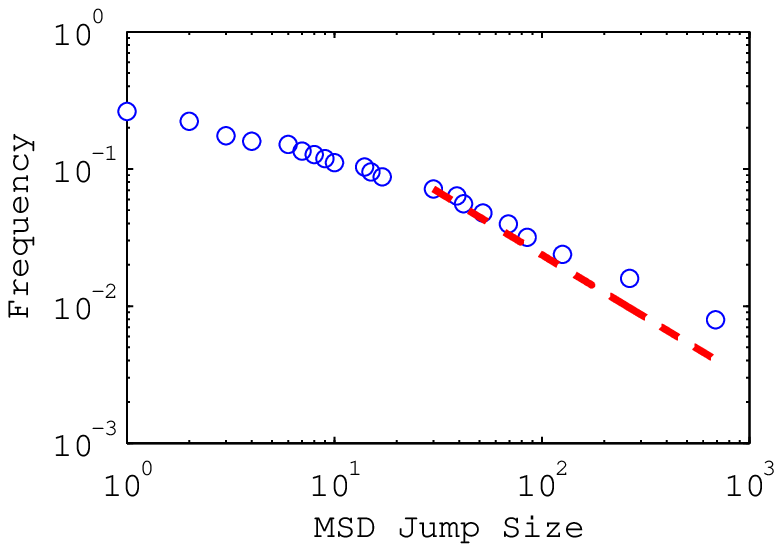}}%
\end{tabular}
}
\thesiscaption[Comparison one of the standard PSO implementation.]{Results obtained with the PSO Laboratory program and the standard PSO using the Rastrigin fitness function.}{fig:comparison_rastrigin_mine}
\end{figure}

The plot in Figure \ref{fig:comp_1_mean_fitness} shows the fitness of all PSO runs averaged in each iteration. This is also the plot which relates the best with the results Lovbkerg \etal obtained. In their research, specific values are not mentioned, but the plot shows not only a similar final fitness, but a similar form overall. The second plot, Figure \ref{fig:comp_1_all_fitness}, shows the fitness obtained by every PSO in each iteration. This plot allows to see the variance of the results and relates to the quality of the solution. If the fitness lines are far and spread apart, the average solution might be lacking in quality. Tighter and more clumped up fitness lines means there is less uncertainty associated with any single solution found. The third plot, Figure \ref{fig:comp_1_mean_msd}, shows the average mean square distance (MSD) from all particles to the centroid of the swarm in every iteration. This plot is important as it related to the dynamic behaviour of the swarm: It shows if the 
particles are clumping up, with a static constant MSD (stagnating), with a constant growing MSD (diverging) or spreading apart. The effect the parameters have in finding a solution can be better understood from this graph, as it will be shown in later sections. 

The last plot, shown in Figure \ref{fig:comp_1_power-law}, is of crucial importance to this project. This shows in more detail some of the characteristics of the average MSD plot. The frequency of the positive increments in the MSD is recorded in fixed size bins. The frequencies are then sorted and normalized into the range $[0, 1]$, where the highest frequency is associated with the value $1$, and the lowest with $0$. The plot shows these frequencies (with dots) against the size of the positive increments in a log-log scale. Additionally, a power-law curve is shown as a dashed line. This line represents the best approximation found for a power-law distribution using maximum likelihood estimates (MLE). Besides the exponent parameter $\alpha$, also the parameter $x_{min}$ is calculated. The fitted curve tries to find a value for $\alpha$ and $x_{min}$ such that the resulting power-law fits as much data as possible.

The bin size used for creating the power-law plot in Figure \ref{fig:comp_1_power-law} is $0.2$ and the interval being considered is $[0, 25]$, which amounts for 50 equally sized bins. This interval is selected as there are only positive MSD increments and none bigger than $25$.

\subsection{Second Standard PSO Test}

A second comparison is performed against another test by Lovbkerg \etal using a different objective function. This time the Griewank objective function with 30 dimensions is  tested. Figure \ref{fig:comparison_griewank} shows the results they obtained with the standard PSO as well as two other variants. Once again, only the result marked as ``STD Average Best'' is of interest to this comparison, the other variants can be disregarded. This result illustrates the average fitness over multiple runs of the standard PSO.

\begin{figure}[ht!]
\centering
\fbox {%
\includegraphics[scale=1.2]{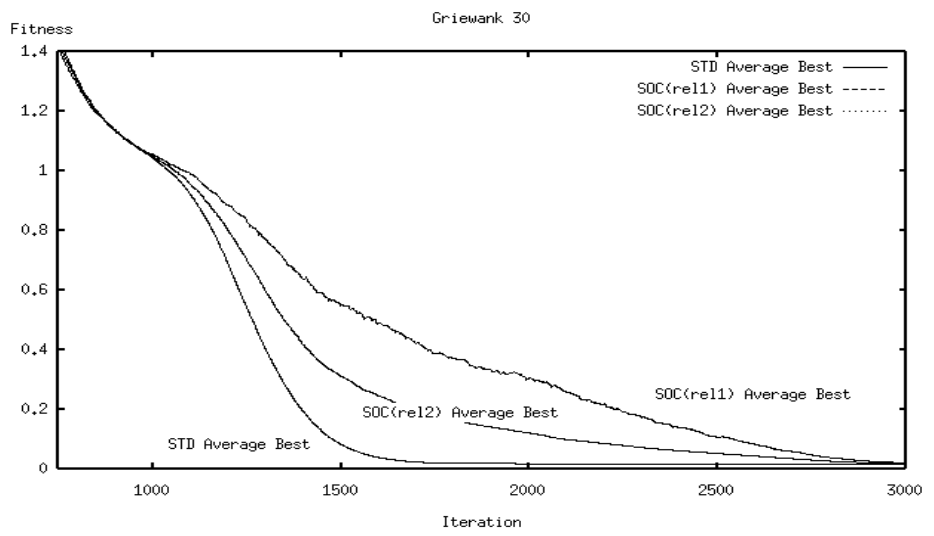}%
}
\thesiscaption[Standard PSO and the Griewank objective function.]{Results obtained by Lovbkerg \etal using the standard PSO with the Griewank objective function.}{fig:comparison_griewank}
\end{figure}

The results of the PSO Laboratory program running the standard PSO with the Griewank objective function are shown in Figure \ref{fig:comparison_griewank_mine}. The plots used to describe the results are the same as the ones used in the previous comparison. The parameters of the PSO where replicated as much as possible from the tests performed by Lovbkerg \etal. 

\begin{figure}[ht!]
\centering
\fbox {%
\begin{tabular}{cc}
\subfloat[Mean fitness of PSOs.]{\label{fig:comp_2_mean_fitness}\includegraphics[scale=0.70]{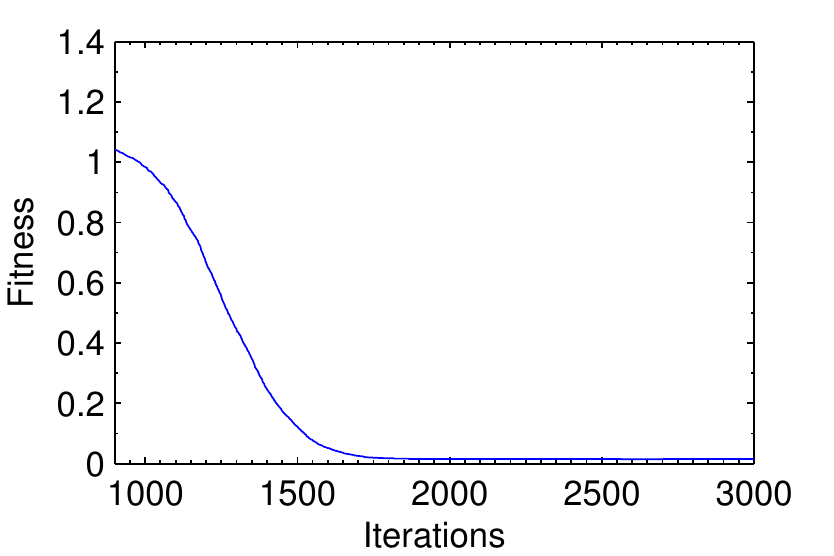}}%
&
\subfloat[Individual fitness of all PSOs.]{\label{fig:comp_2_all_fitness}\includegraphics[scale=0.70]{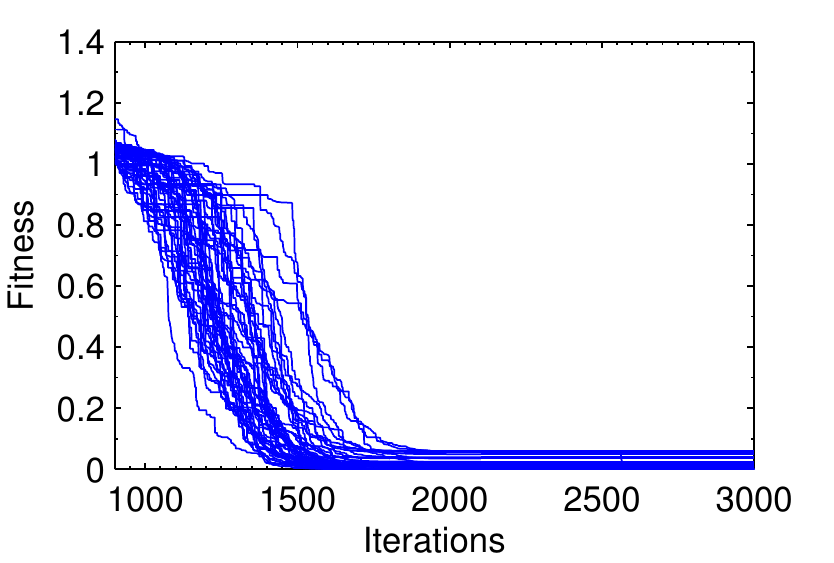}}\\
\subfloat[Mean MSD of all runs.]{\label{fig:comp_2_mean_msd}\includegraphics[scale=0.70]{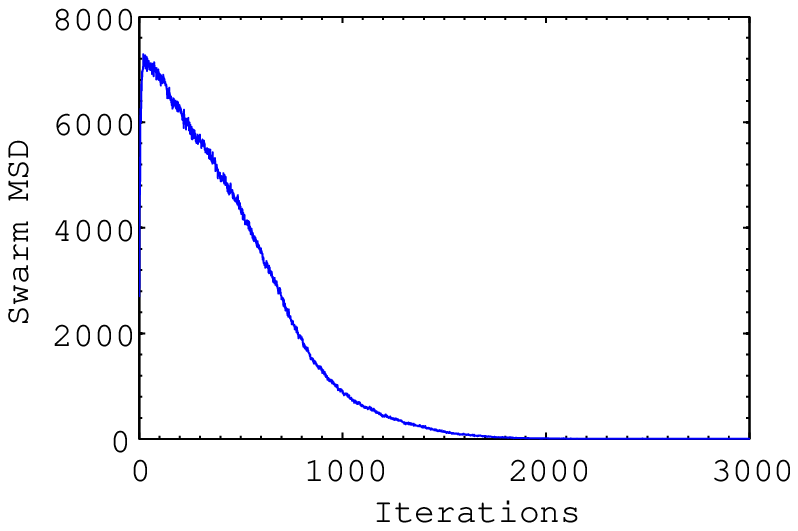}}%
&
\subfloat[Power-law approximation.]{\label{fig:comp_2_power-law}\includegraphics[scale=0.70]{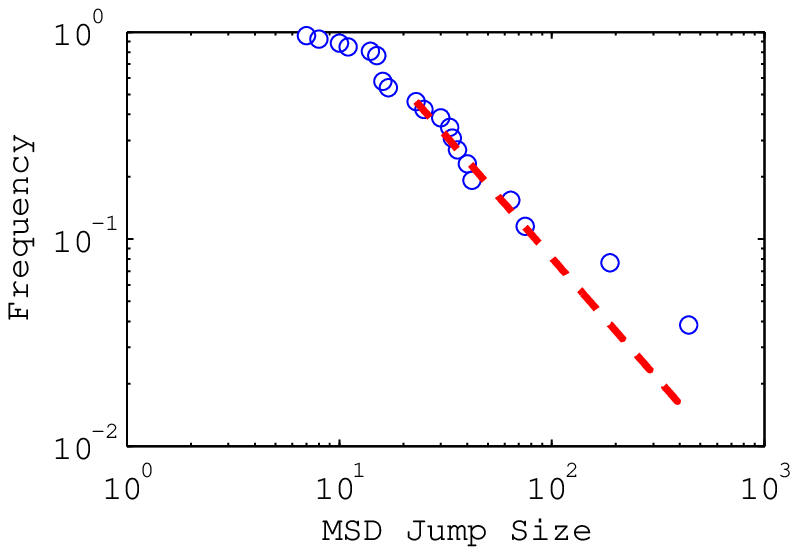}}%
\end{tabular}
}
\thesiscaption[Comparison two of the standard PSO implementation.]{Results obtained with the PSO Laboratory program using the standard PSO and the Griewank fitness function.}{fig:comparison_griewank_mine}
\end{figure}

\subsection{Comparison Results}

It was possible to replicate the results obtained by Lovbkerg \etal with the standard PSO algorithm. In both tests, the final solution as well as the form of the fitness curve, over multiple iterations, was similar. The tests using the Rastrigin function, illustrated in Figures \ref{fig:comparison_rastrigin} and \ref{fig:comp_1_mean_fitness}, have a final average fitness very close to $40$. It can also be appreciated how the solution does not improve much after $2,000$ iterations. The second comparison, illustrated with Figures \ref{fig:comparison_griewank} and \ref{fig:comp_2_mean_fitness}, yield the same conclusions as before. In both plots the final average fitness is close to $0.01$ and the solution stops improving substantially somewhere around iteration $1,500$. As it was possible to replicate the results obtained in both tests, the algorithm is deemed to be working correctly and as expected.

\subsection{Information Beyond Fitness}

To compare results with Lovbkerg \etal only the average fitness graph would be required; nevertheless, a lot information regarding the behaviour of the swarm is lost in such graphs. To understand the results of this research and the effects of adding self-organized criticality to PSOs, the other three plots must be taken into account.

Plot ``b'' in both Figure \ref{fig:comparison_griewank_mine} and \ref{fig:comparison_rastrigin_mine} show the quality of the solution or, in other words, the standard deviation between solutions found over multiple runs of the same algorithm. In the Rastrigin test, the standard PSO algorithm is lacking in quality. From the plot, it can be seen how the solutions of multiple runs are spread apart from each other. The standard deviation is $8.5476$. The solutions found over the Griewank function have better quality, with a standard deviation of $0.0175$.

Plot ``c'' in both tests show that the particles start exploring far away from each other and begin to clump up until the algorithm becomes stagnant. There is a clear point where the particles collapse and stop exploring; the balance between exploration and exploitation is lost. Self-organized criticality, as will be demonstrated in later sections, confer the algorithm with the capacity to keep exploring even after facing stagnation.

Plot ``d'' shows two things at the same time. On the one hand, the frequency of the positive MSD increments is shown with dots. On the other hand, the power-law which best describes the frequencies (calculated with MLE) is shown with a dashed line. These two things allows to detect if a PSO is potentially behaving in a critical manner.

\section{Performance of the Eigencritical Particle Swarm}

The Eigencritical PSO variant has some interesting particularities when evaluated over multiple and single runs. After few iterations the variant immediately stagnates. All particles clump up and do not move much. The particles are specially attracted to the origin, which means that objective functions with their global minimum at the origin are optimized deceptively. For this reason the Schwefel objective function is selected to test the performance of this variant.

Originally, it was expected that parameter selection would have no impact in the performance of this variant; after all, the parameters shape the transformation matrix which is later modified, disregarding the parameters. In practice, nonetheless, parameter selection still plays an important role in this variant. The parameters are not selected to achieve better performance, but rather to avoid the variant from completely diverging or stagnating. If the parameters are too high, the particles continuously fly away from each other; if the parameters are too low, all particles remain static at a single point. It is possible to artificially solve the problem of divergence by setting hard limits to the velocity and position of the particles. This, however, does not solve the problem of stagnation. To test which parameters are suitable with the Schwefel objective function no velocity or boundary limits are set.

The procedure to find suitable parameters is as follows. In the PSO Laboratory program an Eigencritical PSO is set to run for million of iterations with $\alpha_1 = 0.1$, $\alpha_2 = 0.1$ and $\omega = 0.1$. The program allows to change the parameters of the PSO on-the-fly; therefore, every few iterations the parameters are gradually incremented until complete divergence is observed. There is a particular range of parameters where the MSD of the swarm drastically increases after which it collapses again, repetitively. When the right parameters are set, this behaviour is observed throughout the entire run of the algorithm. This allows the swarm to keep exploring for the entire run of the algorithm. This behaviour can be appreciated in Figure \ref{fig:eigencritical_behaviour}.

\begin{figure} [ht]
\centering
\fbox{
  \subfloat[Fitness of the PSO]{\label{fig:eigen_single_fitness}\includegraphics[scale=0.70]{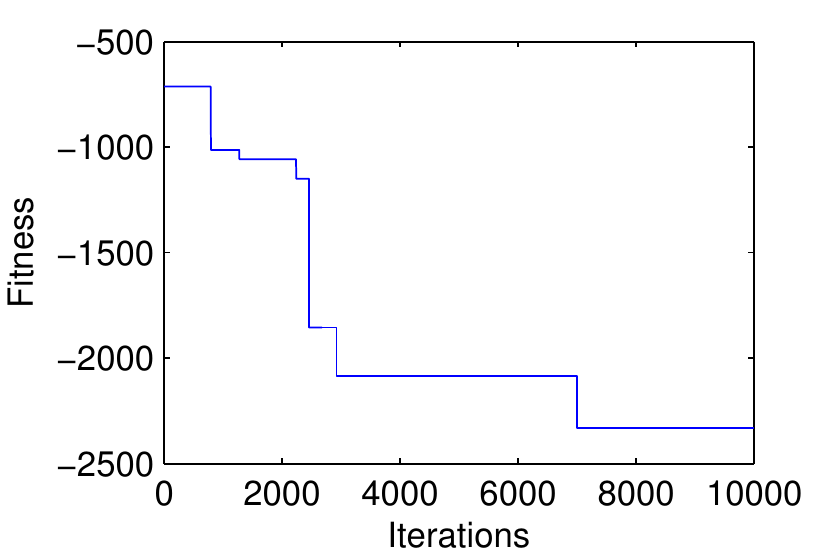}}
  \hspace{8mm}
  \subfloat[MSD of the PSO]{\label{fig:eigen_single_msd}\includegraphics[scale=0.70]{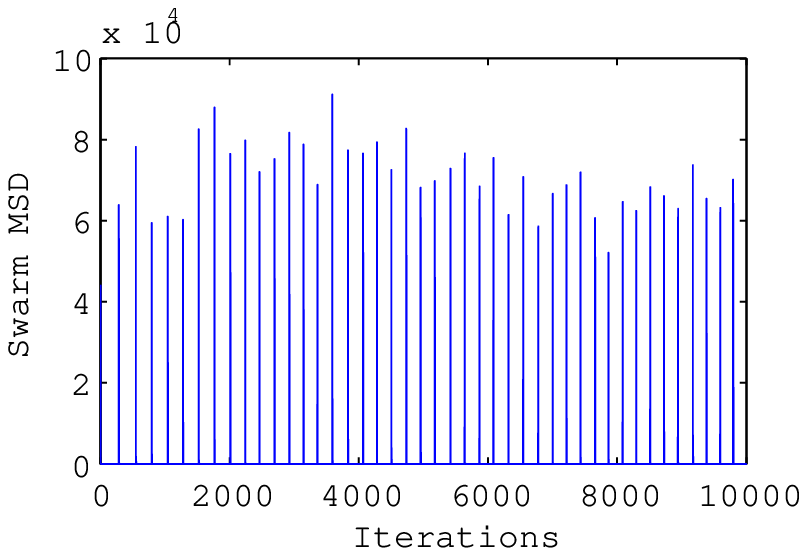}}
}
\thesiscaption[Eigencritical PSO with the right set of parameters.]{Behaviour of the Eigencritical PSO with the right set of parameters.}{fig:eigencritical_behaviour}
\end{figure}

The following tests were performed with the parameter values $\alpha_1 = 0.6$, $\alpha_2 = 0.6$ and $\omega = 0.6$; with 20 particles and 20 dimensions. There are no boundaries set for either the velocity or the position of the particles, and they are evenly initialized inside a hypersphere of radius $500$. Figure \ref{fig:eigen_performance} shows the average results of 50 different runs.

\begin{figure}[ht!]
\centering
\fbox {%
\begin{tabular}{cc}
\subfloat[Mean fitness of PSOs.]{\label{fig:eigen_mean_fitness}\includegraphics[scale=0.70]{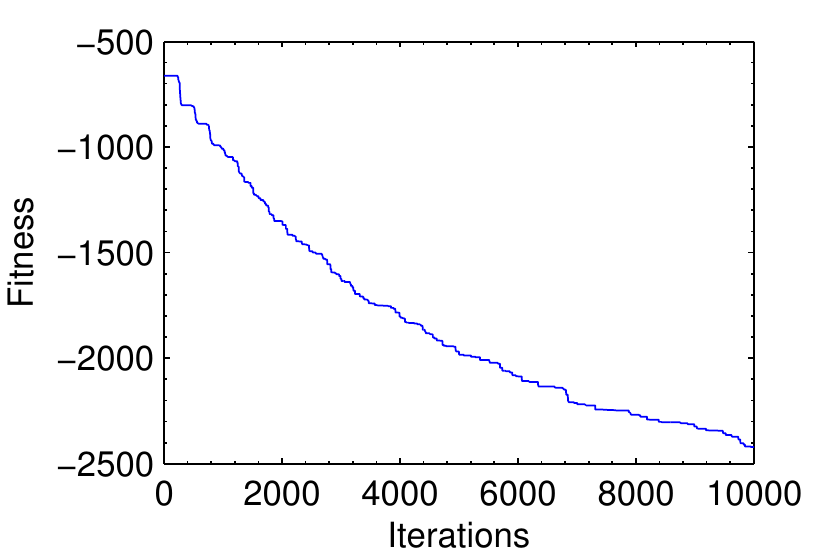}}%
&
\subfloat[Individual fitness of all PSOs.]{\label{fig:eigen_all_fitness}\includegraphics[scale=0.70]{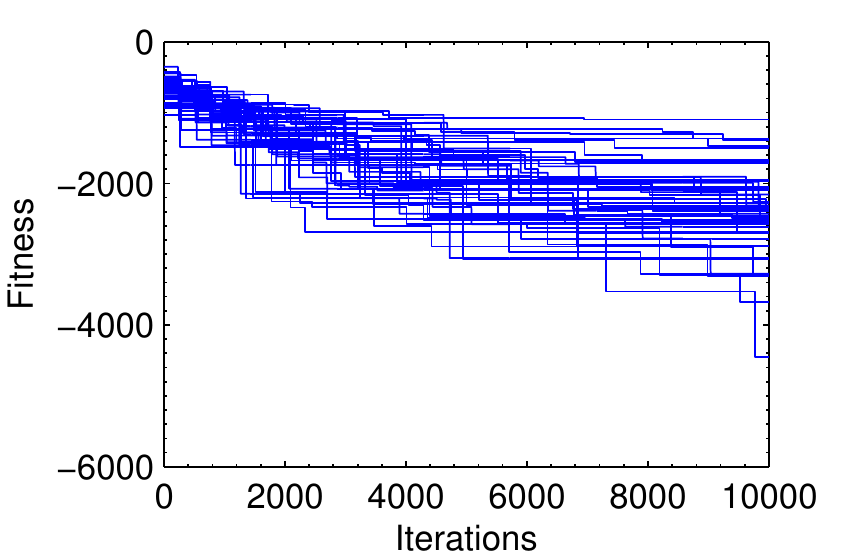}}\\
\subfloat[Mean MSD of all runs.]{\label{fig:eigen_mean_msd}\includegraphics[scale=0.70]{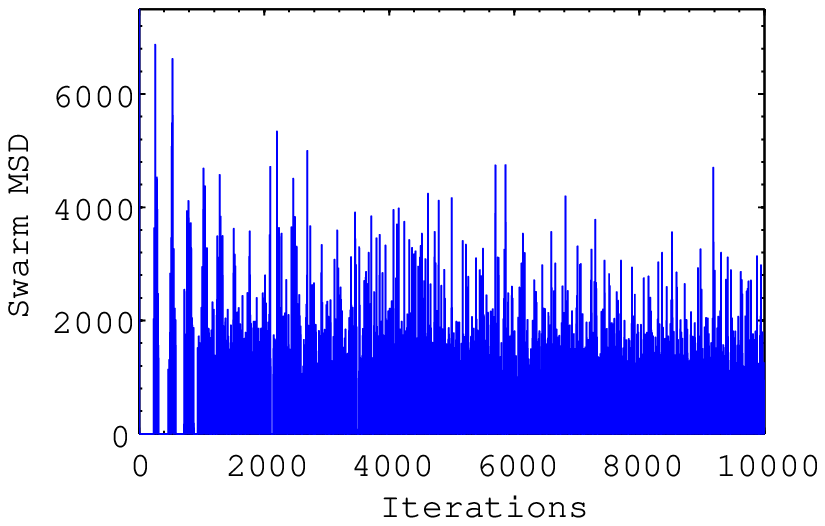}}%
&
\subfloat[Power-law approximation.]{\label{fig:eigen_power-law}\includegraphics[scale=0.70]{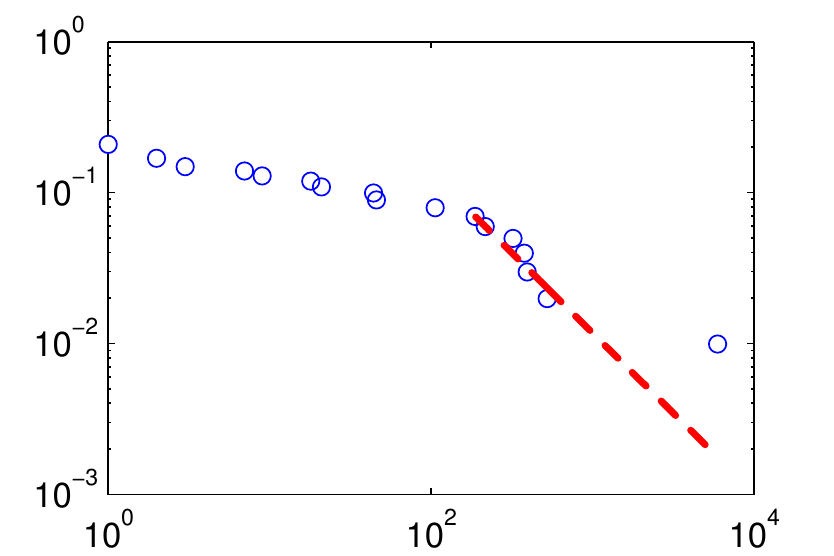}}%
\end{tabular}
}
\thesiscaption[]{Performance results of the Eigencritical PSO variant.}{fig:eigen_performance}
\end{figure}

The performance of the algorithm, in contrast with the standard PSO, is poor. The global optimum of the Schwefel objective function with $20$ dimensions is $-8,379.658$. The eigencritical PSO had an average best fitness of $-2,421.4$ with a standard deviation of $634.96$. The overall best fitness achieved by any single run was $-4,451.4$. There are, however, two important observations: The fitness improves monotonically without signs of stagnation; and the algorithm never stops exploring throughout the entire run. Figure \ref{fig:eigen_power-law} shows that the average behaviour of this variant is still not critical (as there is no power-law distribution describing the dynamics).

This variant might not perform better than the standard PSO, but it certainly has some interesting characteristics which will be exploited in the Adaptive PSO variant.

\section{Performance of the Adaptive Particle Swarm}
\label{sec:adaptive_pso_performance}

The Adaptive PSO variant was inspired by the results obtained with the Eigencritical PSO. The previous observations demonstrate the possibility of having a PSO explore continuously throughout the entire run of the algorithm. The Eigencritical PSO variant has some disadvantages which the Adaptive PSO tries to overcome. The former has high computational requirements making the algorithm impractical for many applications. The latter reduces the computational costs by using measurements expected to relate proportionally to the measurements used in the former \cite{Yazdi2007}. The Eigencritical variant does not scale well. Because linear transformations have to be calculated between the present position of the particles and their future position, more particles means more variables to solve.

Once again, parameter selection still has to be performed in some way or form. Just as with the Eigencritical PSO, there are parameter sets which make the algorithm completely divergent or stagnant. The procedure for selecting parameters is similar to the one performed in the previous section. The only difference being that the algorithm has to be run multiple times with each new set of parameters, as this variant does not allow them to be modified on-the-fly. The suitable parameter range, though, is completely different from the working range of the Eigencritical variant. The behaviour of the algorithm outside of this range is also peculiar. Figure \ref{fig:adaptive_param_regions} shows the MSD behaviour of three particle swarms using different sets of parameters. 

Table \ref{table:adaptive_params_used} shows the main PSO parameters used in each swarm. The same metric and parameter modification rule are used. The modification rule is the average velocity norm of all particles, with the \textit{dependant rule}. The \textit{adaptive epsilon} is set to $\varepsilon = 0.1$. The Schwefel function is used as objective function.

\begin{figure} [ht]
\centering
\fbox{
  \subfloat[Parameters to the left of the ideal range.]{\label{fig:adaptive_single_left}\includegraphics[scale=0.5]{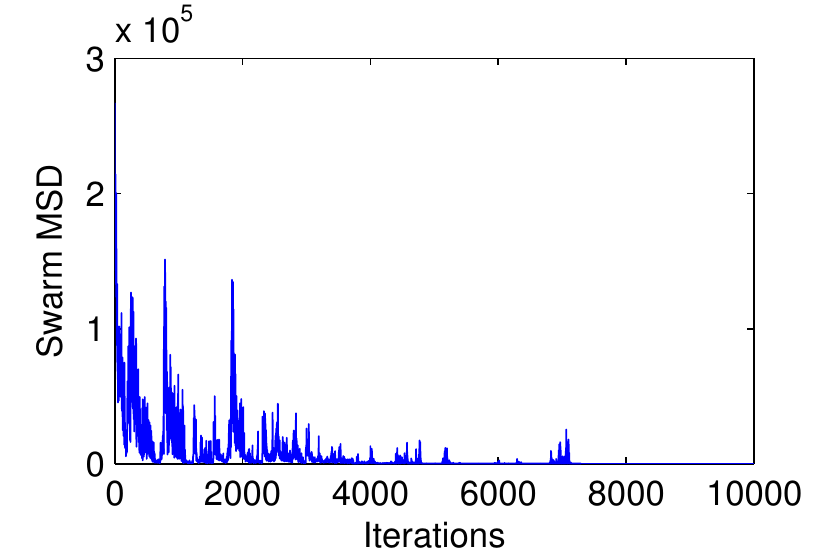}}%
  \hspace{2mm}
  \subfloat[Parameters within the ideal range.]{\label{fig:adaptive_single_within}\includegraphics[scale=0.5]{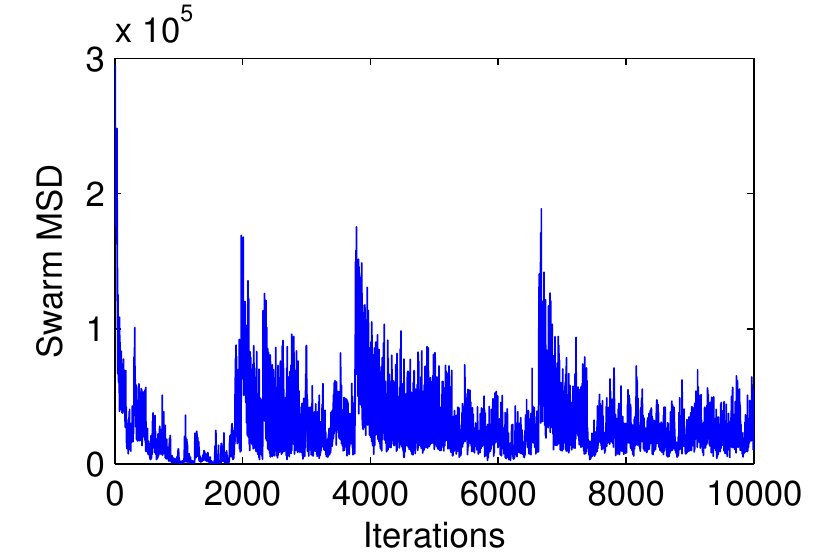}}%
  \hspace{2mm}
  \subfloat[Parameters to the right of the ideal range.]{\label{fig:adaptive_single_right}\includegraphics[scale=0.5]{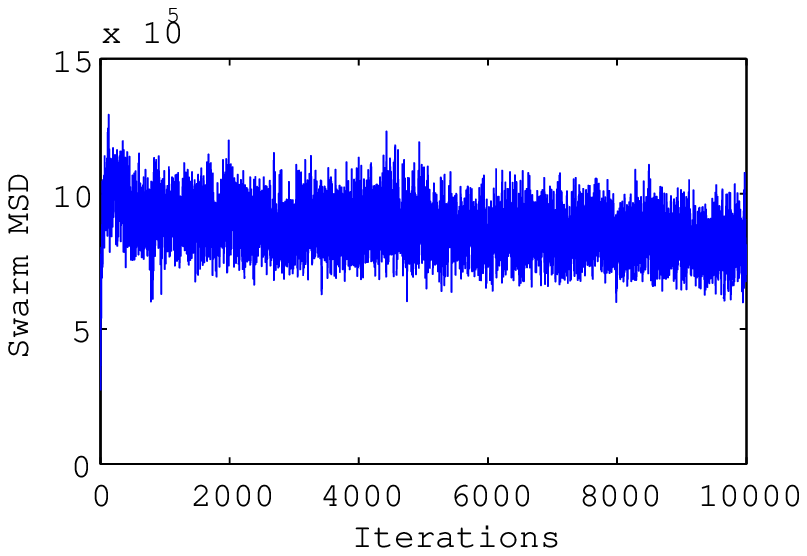}}%
}
\thesiscaption[MSD of the Adaptive PSO and parameter selection.]{MSD of the Adaptive PSO depending on the starting parameters.}{fig:adaptive_param_regions}
\end{figure}
\begin{table}[ht]
\centering    
\begin{tabular}{| c | c | c | c |} \hline
\textbf{Figure} & $\mathbf{\alpha_1}$ & $\mathbf{\alpha_2}$ & $\mathbf{\omega}$   \\ \hline
Figure \ref{fig:adaptive_single_left} &1 & 1 & 0.800 \\
Figure \ref{fig:adaptive_single_within} &1 & 1 & 0.815 \\
Figure \ref{fig:adaptive_single_right} &1 & 1 & 0.900 \\ \hline
\end{tabular}
\caption{Recommended parameter values for balancing exploitation and exploration.}
\label{table:adaptive_params_used}
\end{table}

Even though only the parameter $\omega$ was modified in the previous test, modifying the parameters $\alpha_1$ and $\alpha_2$ impact the behaviour of the algorithm just as well; however, $\omega$ has the strongest influence. As can be appreciated from the $\omega$ values in Table \ref{table:adaptive_params_used}, small modifications are all it takes to observe big changes in the exploratory capabilities of the algorithm. Originally, this variant used the computed metric of the swarm to directly modify the parameters (with the modification rule selected by the user). Because of the high sensitivity to the modifications, the sigmoid-like function (\ref{eq:adaptive_sigmoid}) was introduced to avoid complete divergence in the algorithm.

\begin{figure}[ht!]
\centering
\fbox {%
\begin{tabular}{cc}
\subfloat[Mean fitness of PSOs.]{\label{fig:adaptive_schwefel_mean_fitness}\includegraphics[scale=0.70]{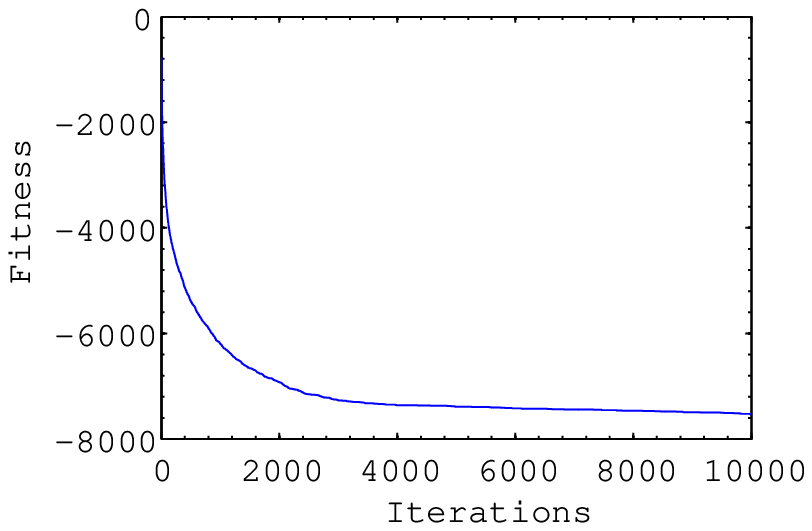}}%
&
\subfloat[Individual fitness of all PSOs.]{\label{fig:adaptive_schwefel_all_fitness}\includegraphics[scale=0.70]{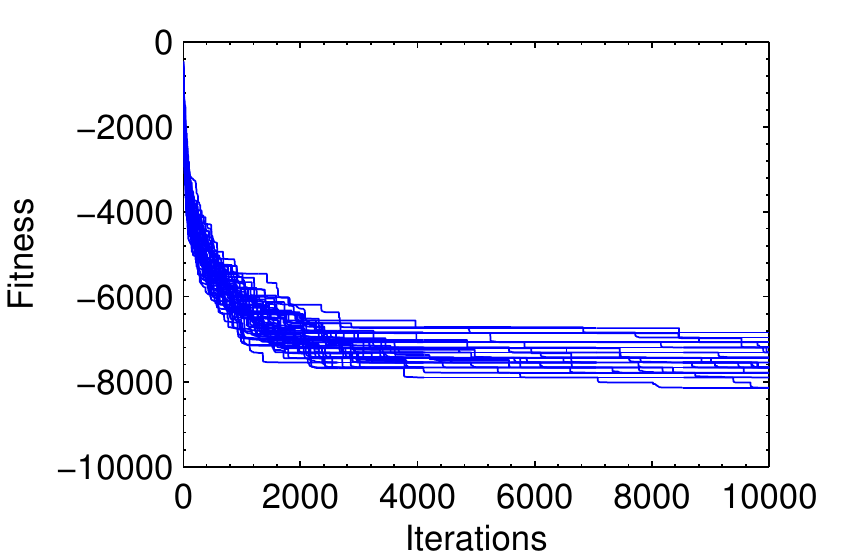}}\\
\subfloat[Mean MSD of all runs.]{\label{fig:adaptive_schwefel_mean_msd}\includegraphics[scale=0.70]{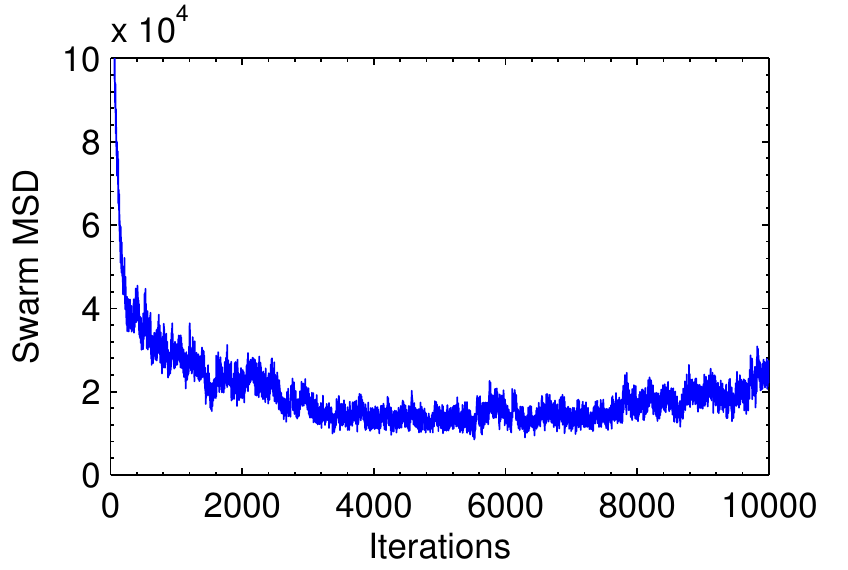}}%
&
\subfloat[Power-law approximation.]{\label{fig:adaptive_schwefel_power-law}\includegraphics[scale=0.70]{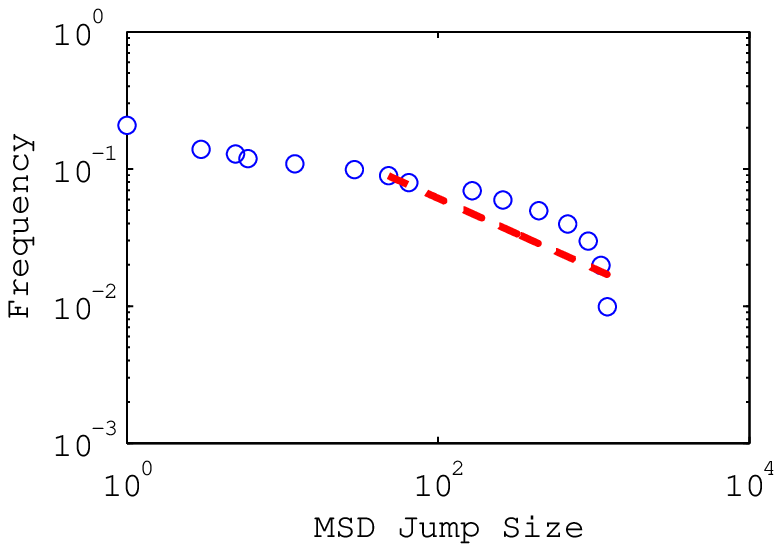}}%
\end{tabular}
}
\thesiscaption[Performance of Adaptive PSO (Schwefel).]{Performance results of the Adaptive PSO variant with the Schwefel objective function.}{fig:adaptive_performance_schwefel}
\end{figure}

To test the performance of the Adaptive PSO, the parameters in the second row of Table \ref{table:adaptive_params_used} are used along with all other parameters previously mentioned. The tested objective function is still the Schwefel function and the algorithm is executed 50 times for statistical purposes. The results obtained are shown in Figure \ref{fig:adaptive_performance_schwefel}. These results are promising as the quality of the solution and the best average fitness have improved dramatically compared to the results of the Eigencritical PSO variant, shown in Figure \ref{fig:eigen_performance}. The best average fitness obtained was $-7,528.6$ with a standard deviation of $284.95$. It is also worth noting that this performance is similar to the standard PSO with hand picked parameters\footnote{The parameters used for testing the standard PSO which gave the best results were $\alpha_1 = 2$, $\alpha_2 = 2$ and $\omega = 0.8$. These parameters were selected by trial and error.}, where the best average 
fitness was $-7,396.6$ with a standard deviation of $246.64$.

The main advantage of this PSO variant lies in its ability to continuously explore. If the algorithm is executed for a long time, as shown in Figure \ref{fig:adaptive_keeps_improving}, the solution keeps improving. After $100,000$ iterations, the best average fitness is $8,212.2$ with a standard deviation of $240.34$; which is an excellent result given the deceptive characteristics of the objective function.

\begin{figure}[ht!]
\centering
\fbox {%
\includegraphics[scale=1]{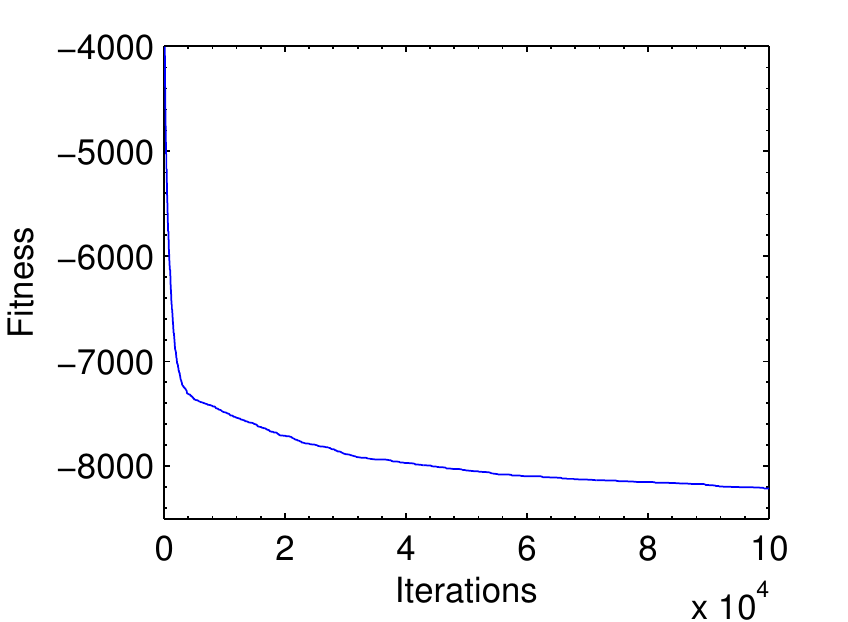}%
}
\thesiscaption[Adaptive PSO fitness in the long run.]{The best average fitness of the Adaptive PSO keeps improving in the long run.}{fig:adaptive_keeps_improving}
\end{figure}

\section{Is Self-Organized Criticality Present?}

\begin{figure} [ht]
\centering
\fbox{
  \begin{tabular}{cc}
  \subfloat[Fitness of an Adaptive PSO run..]{\label{fig:critical_swarm_fitness}\includegraphics[scale=0.70]{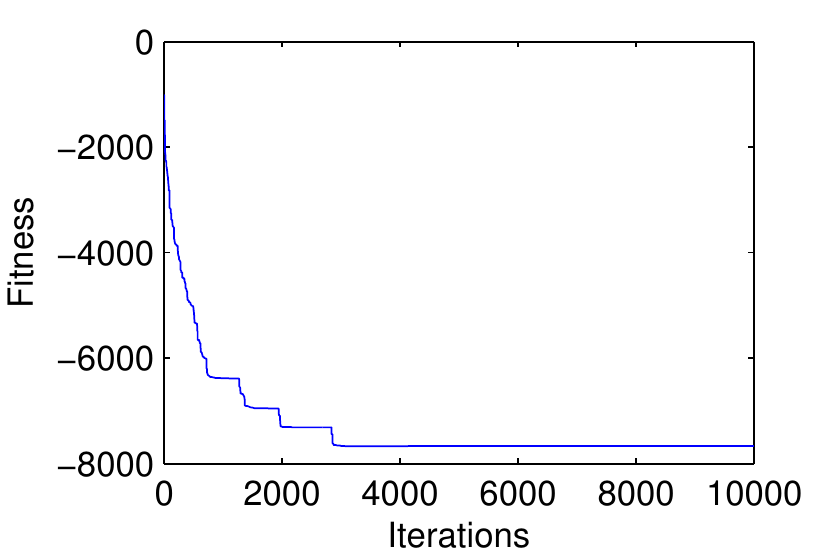}}%
  &
  \subfloat[MSD of an Adaptive PSO run.]{\label{fig:critical_swarm_msd}\includegraphics[scale=0.70]{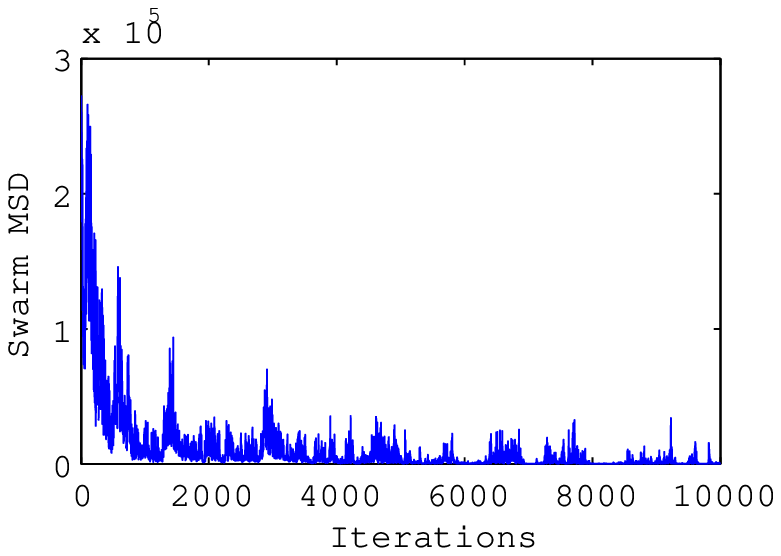}}\\
  \subfloat[Interpolation of the data points.]{\label{fig:critical_swarm_power-law_1}\includegraphics[scale=0.70]{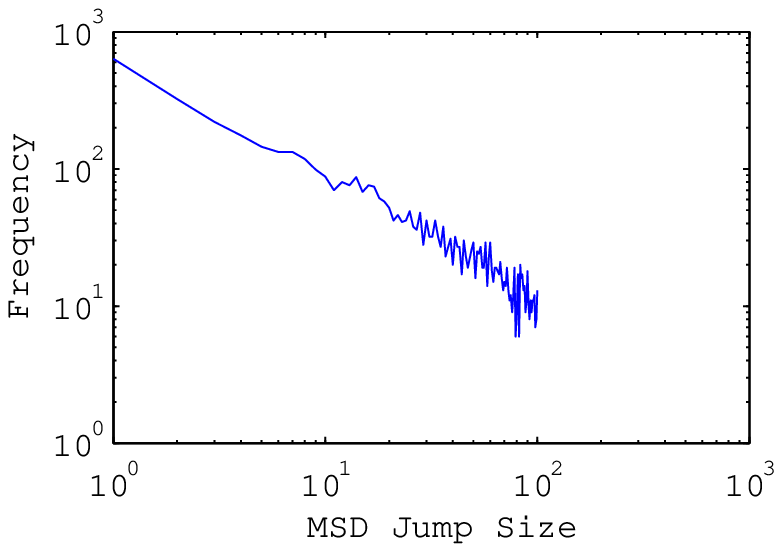}}%
  &
  \subfloat[Power-law approximation.]{\label{fig:critical_swarm_power-law_2}\includegraphics[scale=0.70]{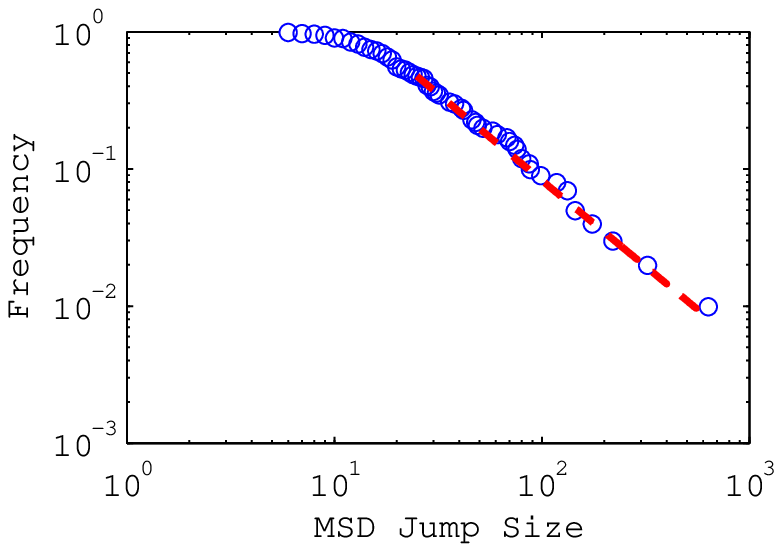}}%
  \end{tabular}
}
\thesiscaption[Analysis of a single Adaptive PSO run.]{Analysis of a single Adaptive PSO run using parameters within the range of suitable parameters.}{fig:critical_swarm}
\end{figure}

Previous results have analysed the average MSD of multiple PSO runs. This average MSD has not shown any interesting properties related to criticality and power-laws. If the focus is shifted towards individual PSO runs, however, interesting results are observed. Figure \ref{fig:critical_swarm} shows the results of a single PSO run. In particular, Figure \ref{fig:critical_swarm_power-law_1} and \ref{fig:critical_swarm_power-law_2} suggest the presence of a critical swarm. The first of these two figures shows a log-log scale plot where the data pertaining to the frequency and size of the MSD jumps is interpolated to show lines. The frequencies are not normalized, as is the case with the second figure. The frequency distribution closely resembles a power-law distribution. Similar results are observed in multiple PSO runs.

\section{Other Metrics and Rules}

Section \ref{sec:adaptive_pso} introduced different metrics for measuring the stagnation/divergence behaviour of a PSO, and different rules to modify the parameters given a metric. Previous sections have only used the \textit{average velocity norm metric} with the \textit{dependant rule}. The combination of these two had several benefits which other combinations did not exhibit: 
\begin{enumerate}
\item The parameters of the Adaptive PSO variant are easy to tune. In particular, the \textit{adaptive epsilon} parameter is less sensitive to small changes.
\item As the \textit{average velocity norm metric} is using the normalized velocity, the range of the metric is constrained. The modification rule is less likely to abruptly (and incorrectly) modify the parameters.
\item The best average fitness and quality of solution are better than with any other metric-rule pairs.
\item It appears that self-organized criticality is spotted in the dynamic behaviour of individual swarms.
\end{enumerate}
It is possible to find a suitable set of parameters for which other metric-rule pairs perform well; however, power-laws are not spotted.

These observations suggest that the average velocity norm of all particles relates well to the stagnation/divergence behaviour of the swarm. If the average velocity increases from one iteration to another, the algorithm is in the process of diverging; on the other hand, if the average velocity decreases the particles are stopping and, therefore, stagnating. By increasing the parameters when the swarm is observed to be stagnating, the particles are forced to move farther distances. The opposite happens when the particles diverge. 

The \textit{dependant rule} for modifying the parameters seems to perform well because it causes the standard PSO parameters ($\alpha_1$, $\alpha_2$ and $\omega$) to balance each other out. This is also the reason why the initial values of the parameters impact the performance of the algorithm. If the parameters are stretching or compressing too much, the algorithm either diverges or converges. Good results are observed when the parameters are fluctuating between $\pm0.4$ throughout all iterations.

  \chapter{Conclusion}

Heuristics study methods and techniques for discovering solutions and solving problems \cite{Zanakis1981}; these are usually simple, more than often inspired by common sense and natural phenomena. Their purpose is not to find optimal solutions but, rather, a good enough solution given time, memory and information constraints. Self-organized criticality has been observed in many natural and man-made phenomena. It was expected for the concept to find some room in the filed of heuristic algorithms, specially in those inspired by nature.

Particle Swarm Optimization (PSO) is a nature-inspired heuristic algorithm which, given its characteristics, was selected for experimentation with the concept of self-organized criticality. The objectives of this research were, one, to find ways to confer particle swarms with self-organized criticality and, two, to test if a critical swarm exhibits useful behaviours in the process of exploiting and exploring the search space.

Heuristic algorithms in general need to balance exploitation and exploration to achieve good results. An algorithm which only exploits will easily get trapped in local optima; an algorithm which only explores will achieve inadequate or poor results more than often. Self-organized criticality is used to balance these two concepts with positive results. A particle swarm variant displaying self-organized criticality was able to be developed. This variant, along with another non-critical variant, were tested against the standard particle swarm in search for improvements.

The first variant was called Eigencritical PSO. Although this variant did not exhibit a critical behaviour, it laid down the fundamentals for a second variant which would possess such characteristic. The Eigencritical PSO used concepts acquired from interpreting particle swarms as dynamic systems. It was demonstrated how modifying the standard PSO parameters is all it takes to drive the swarm towards a middle state where the algorithm neither diverges nor converges: a critical state. From the experiments, the Eigencritical PSO was observed to stagnate immediately; however, every once in a while large chain reactions in the movement of the particles would occur which would make the swarm spread apart and cover large areas, just to return to its stagnant state again. This is a interesting and desirable behaviour from which particle swarms could benefit.

The Eigencritical PSO, besides not being critical at all, had some other disadvantages which the next variation tried to overcome. The Adaptive PSO variation builds on top of what was observed in the Eigencritical PSO experiments. It also tried to avoid the expensive calculations performed by the Eigencritical PSO. Instead of having to compute a linear transformation of particle positions from the present state to a future state and obtain the largest eigenvalue of such transformation, simple and easy to calculate metrics where used to determine if the swarm was converging or diverging. Depending on these criteria, the Adaptive PSO variant adaptively modifies the parameters of the algorithm to achieve a state of neither convergence nor divergence. 

The Adaptive PSO variant showed promising results in the long run and good results in the short run. All tests were performed with the Schwefel objective function as it is not only difficult to optimize, but deceptive as well. After $10,000$ iterations the solutions found are close to the ones obtained by the standard PSO. The advantages of this variant, however, are better observed in the long run and whenever complicated or deceptive functions are used. Because exploration never ceases, better solutions are found as the algorithm is able to escapes local optima easily while exploring far away regions. Self-organized criticality balances the exploratory capabilities of the swarm with a power-law distribution: regions are explored inversely proportional to their size. Most of the time particles tend to stay together exploring small regions; there are moments, nevertheless, where the swarm explodes in size, explores big regions and returns to its starting size. After $100,000$ iterations the average solution found and its quality is considerably better than the one obtained by the standard PSO.

Self-organized criticality is said to be responsible for balancing the exploratory capabilities of the Adaptive PSO variant as it appears to make itself present in the distribution of the swarm size and the frequency of its growth. Every iteration the mean square distance (MSD) between every particle and the centroid of the swarm is recorded. If the difference between two consecutive MSD records is plotted in a log-log scale graph for individual PSO runs, what appears to be a straight line with negative slope is observed. Although this is evidence of the existence of a power-law distribution, self-organized criticality might not necessarily be present. The claim of self-organized criticality is further supported by the presence of sub-critical and super-critical regions in the parameter space. Certain combinations of parameters make the swarm immediately stagnate (sub-critical region) or diverge (super-critical region); there is a specific range in the middle which allows the swarm to behave in what appears to be a critical fashion.

\section{Final Remarks and Observations}

A PSO variant able to perform parameter selection using self-organized criticality was able to be created; nevertheless, it is still not fully independent of user interaction. It was hoped that parameter selection would be completely integrated into the particle swarm heuristic itself, but the user is still required to tune some parameters by hand. In the process of achieving this objective it was possible to observe the benefits of self-organized criticality and the balance it brings to the exploratory capabilities of PSOs.

Even though users are still required to tune parameters to achieving good results, parameter selection becomes easier to perform through trial and error as there is a well-defined objective to parameter tuning with intuitive feedback. The usual approach to parameter selection in the standard PSO involves selecting a set of parameters after which the algorithm is executed and the results recorded. This needs to be performed multiple times until it is made clear which parameters yield the best results. In contrast, parameter selection in the Adaptive PSO variant becomes the task of finding a set of parameters which neither makes the algorithm stagnate nor diverge. Plotting the MSD of the swarm against the number of iteration is of great aid to perform this. After setting an initial and arbitrary set of parameters, the swarm is executed for a small fraction of the desired iterations. If the MSD of the swarm approaches zero rapidly, the initial parameters need to be increased. If, on the other hand, the MSD constantly increases the initial parameters need to be decreased. 

\section{Future Work}

Plenty of things can still be done to follow up what has been researched in this project. The most interesting avenue of research involves exploring the Adaptive PSO variant in more detail. In this research three different metrics for assessing the (divergent or stagnant) state of a particle swarm were examined. Two rules for modifying the parameters using these metrics were proposed. Metrics can be easily defined, but the modification rules are not trivial to develop. In this research both proposed rules modify the parameters in one direction: in every iteration all parameters are either increased or decreased in size. Further research could explore the creation of rules which could modify each parameter independently, or even create ``meta-rules'' which could combine different simple rules and metrics together.

The Adaptive PSO variant introduced the \textit{adaptive epsilon} parameter. In theory it should be possible to modify this parameter throughout a PSO run by decreasing or increasing it every iteration. Early in the execution of the algorithm small \textit{adaptive epsilon} values could help the parameters settle into a suitable range; afterwards, large values would allow better exploration of the parameter space. This new parameter could be modified adaptively either linearly or in a sinusoidal pattern in proportion to the total number of iterations.


  \bibliography{thebibliography}{}
  \bibliographystyle{apalike}
\end{document}